%% file: main.tex
\documentclass[10pt,journal,compsoc]{IEEEtran}
\ifCLASSOPTIONcompsoc
  \usepackage[nocompress]{cite}
\else
  \usepackage{cite}
\fi
\usepackage{soul}
\usepackage{url}
\usepackage[hidelinks]{hyperref}
\usepackage[utf8]{inputenc}
\usepackage{graphicx}
\usepackage{amsmath}
\usepackage{tabu}
\usepackage{amsthm}
\usepackage{booktabs}
\usepackage{algorithm}
\usepackage{algorithmic}
\usepackage{subfigure}
\usepackage{xspace}
\usepackage[dvipsnames]{xcolor}
\usepackage{multirow}
\usepackage{enumitem}
\usepackage{bm}

\usepackage{caption}
\usepackage{amsmath,amsfonts,amssymb}
\usepackage{listings}
\usepackage{makecell}
\usepackage[figuresright]{rotating}
\usepackage{hyperref}
\usepackage[edges]{forest}
\hypersetup{
    urlcolor=blue
}

\newcommand{\paratitle}[1]{\vspace{1.0ex}\noindent\textbf{#1}}
\newcommand{\ie}{\emph{i.e.,}\xspace}

\newcommand{\eg}{\emph{e.g.,}\xspace}
\newcommand{\etal}{\emph{et al.}\xspace}
\newcommand{\ignore}[1]{}

\newcommand{\citeal}[1]{\citeauthor{#1}~[\citeyear{#1}]}

\newcommand{\tabincell}[2]{\begin{tabular}{@{}#1@{}}#2\end{tabular}}
\newcommand{\jhadd}[1]{\textcolor{magenta}{#1}}
\newcommand{\jhcomment}[1]{\textcolor{magenta}{[JH: #1]}}
\newcommand{\yscomment}[1]{\textcolor{orange}{[YS: #1]}}
\newcommand{\ysadd}[1]{\textcolor{orange}{#1}}
\newcommand{\glcomment}[1]{\textcolor{red}{[GL: #1]}}
\newcommand{\gladd}[1]{\textcolor{red}{#1}}
\renewcommand{\figurename}{FIGURE}

\ifCLASSINFOpdf
\else
\fi
\begin{document}
%
\title{Complex Knowledge Base Question Answering: A Survey}
%
%
%
%

\author{Yunshi~Lan*, Gaole~He*, Jinhao~Jiang, Jing~Jiang, Wayne~Xin~Zhao~\IEEEmembership{Member,~IEEE},  and~Ji-Rong~Wen~\IEEEmembership{Senior Member,~IEEE},
\IEEEcompsocitemizethanks{
\IEEEcompsocthanksitem * Y. Lan and G. He contribute equally to this work.
\IEEEcompsocthanksitem Yunshi Lan is with School of Data Science and Engineering, East China Normal University.
E-mail: yslan@dase.ecnu.edu.cn 
\IEEEcompsocthanksitem Jing Jiang is with School of Computing and Information System, Singapore Management University.
E-mail: jingjiang@smu.edu.sg.
\IEEEcompsocthanksitem G.He is with the School of Information, Renmin University of China, and Beijing Key Laboratory of Big Data Management and Analysis Methods.
E-mail: hegaole@ruc.edu.cn
\IEEEcompsocthanksitem W.X. Zhao (corresponding author) , Jinhao Jiang and J. Wen are with Gaoling School of Artificial Intelligence, Renmin University of China, and Beijing Key Laboratory of Big Data Management and Analysis Methods.
E-mail: batmanfly@gmail.com.

}
\thanks{Manuscript revised xxx.}}

%
%

\markboth{Journal of \LaTeX\ Class Files,~Vol.~14, No.~8, August~2015}%
{Shell \MakeLowercase{\textit{et al.}}: Bare Demo of IEEEtran.cls for Computer Society Journals}
%



\IEEEtitleabstractindextext{%
\begin{abstract}

\ignore{
  Knowledge base question answering (KBQA) aims to answer a question over a knowledge base (KB).
  Recently, a great amount of studies focuses on semantically or syntactically complicated questions.
  In this paper, we elaborately summarize the typical challenges and solutions for complex KBQA. 
  We begin with introducing the task  and available datasets.
  Next, we present the two mainstream categories of methods for complex KBQA, namely semantic parsing-based~(SP-based) methods and information retrieval-based~(IR-based) methods.
  We then review the advanced methods comprehensively in the perspective of two categories.
  Specifically, we explicate their solutions to the typical challenges.
  Finally, we conclude and discuss some directions for future research.~\glcomment{Add more content here}  
}
Knowledge base question answering (KBQA) aims to answer a question over a knowledge base (KB). 
Early studies mainly focused on answering simple questions over KBs and achieved great success.
However, their performance on complex questions are still far from satisfaction.
Therefore, in recent years, researchers propose a large number of novel methods, which looked into the challenges of answering complex questions.
In this survey, we review recent advances on KBQA with the focus on solving complex questions, which usually contain multiple subjects, express compound relations, or involve numerical operations. 
In detail, we begin with introducing the complex KBQA task and relevant background.  
Then, we present two mainstream categories of methods for complex KBQA, namely semantic parsing-based~(SP-based) methods and information retrieval-based~(IR-based) methods. 
Specifically, we illustrate their procedures with flow designs and discuss their difference and similarity.
Next, we summarize the challenges that these two categories of methods encounter when answering complex questions, and explicate advanced solutions as well as techniques used in existing work. After that, we discuss the potential impact of pre-trained language models (PLMs) on complex KBQA. To help readers catch up with SOTA methods, we also provide a comprehensive evaluation and resource about complex KBQA task.
Finally, we conclude and discuss several promising directions related to complex KBQA for future research.
\end{abstract}

}

\maketitle

\IEEEdisplaynontitleabstractindextext

%
\IEEEpeerreviewmaketitle

\input{sec-intro}
\input{sec-back}

\input{sec-pre}
\input{sec-main}

\input{sec-other}

\input{sec-eval_resource}
\input{sec-con}


%

\ignore{
\appendices
\section{Proof of the First Zonklar Equation}
Appendix one text goes here.

\section{}
Appendix two text goes here.
}

\ignore{
\ifCLASSOPTIONcompsoc
  \section*{Acknowledgments}
\else
  \section*{Acknowledgment}
\fi

The authors would like to thank...
}

\ifCLASSOPTIONcaptionsoff
  \newpage
\fi



%
\bibliographystyle{IEEEtran}
\bibliography{survey}
\ignore{
\begin{IEEEbiography}[{\includegraphics[width=1in,height=1.25in,clip,keepaspectratio]{Figures/Yunshi_Photo.jpeg}}]{Yunshi Lan} 
obtained her Ph.D.~degree from the School of Computing and Information Systems at Singapore Management University in 2020. 
She is currently an associated professor at School of Data Science and Engineering of East China Normal University.
Her research interests lie in natural language processing with a focus on knowledge base question answering.
\end{IEEEbiography}

\begin{IEEEbiography}[{\includegraphics[width=1in,height=1.25in,clip,keepaspectratio]{Figures/Gaole.jpg}}]{Gaole He} is currently a Ph.D.~student at Delft University of Technology. He received his master degree from the School of Information at Renmin University of China in 2021. His research focuses on graph minging, knowledge representation and knowledge reasoning.
\end{IEEEbiography}

\begin{IEEEbiography}[{\includegraphics[width=1in,height=1.25in,clip,keepaspectratio]{Figures/Jinhao.png}}]{Jinhao Jiang} is currently a Ph.D.~student at Renmin University of China. He received his bachelor degree from school of information and software engineering, University of Electronic Science and Technology of China in 2021. His research focuses on natural language processing, knowledge base and knowledge reasoning.
\end{IEEEbiography}

\begin{IEEEbiography}[{\includegraphics[width=1in,height=1.25in,clip,keepaspectratio]{Figures/JJ_Photo.jpeg}}]{Jing Jiang} 
received her Ph.D.~degree in Computer Science from the University of Illinois at Urbana-Champaign in 2008.
Currently, she is a professor in the School of Computing and Information Systems at the Singapore Management University.
Her research interests include natural language processing, text mining and machine learning.
\end{IEEEbiography}

\begin{IEEEbiography}[{\includegraphics[width=1in,height=1.25in,clip,keepaspectratio]{Figures/xinzhao.jpeg}}]{Wayne Xin Zhao} 
received the Ph.D.~degree from
Peking University in 2014. He is currently a tenured associated professor in Gaoling School of Artificial Intelligence, Renmin University of China. His research interests are web text mining and natural language processing. He has published a number of papers in international conferences and journals such as ACL, SIGIR, SIGKDD, WWW, ACM TOIS, and IEEE TKDE.
He is a member of the IEEE.
\end{IEEEbiography}

\begin{IEEEbiography}[{\includegraphics[width=1in,height=1.25in,clip,keepaspectratio]{Figures/jrwen.jpeg}}]{Ji-Rong Wen} 
received Ph.D.~degree from the Chinese Academy of Science in 1999. He is a professor at the Renmin University of China. He was a senior researcher and research manager with Microsoft Research
from 2000 to 2014. His main research interests include web data management, information retrieval (especially web IR), and data mining. He is a senior member of the IEEE.
\end{IEEEbiography}

%

\begin{IEEEbiography}{Michael Shell}
Biography text here.
\end{IEEEbiography}

\begin{IEEEbiographynophoto}{John Doe}
Biography text here.
\end{IEEEbiographynophoto}


\begin{IEEEbiographynophoto}{Jane Doe}
Biography text here.
\end{IEEEbiographynophoto}  
}





\end{document}

%% file: sec-intro.tex
\IEEEraisesectionheading{\section{Introduction}\label{sec:intro}}

%
%
%
%

\IEEEPARstart{K}{} nowledge base~(KB) is a structured database that contains a collection of facts (alias triples) in the form (\emph{subject}, \emph{relation}, \emph{object}). %
Large-scale KBs, such as Freebase~\cite{Bollacker-SIGMOD-2008}, DBPedia~\cite{dbpedia-2015}, Wikidata~\cite{Thomas-WWW-2016}, and YAGO~\cite{Suchanek-WWW-2007}, have been constructed to serve many downstream tasks. 
Among them, knowledge base question answering~(KBQA) is a task that aims to answer natural language questions with KBs acting as its knowledge source.
Nowadays, KBQA has attracted intensive attention from researchers as it plays an important role in many intelligent applications such as Apple Siri, Microsoft Cortana and so on~\cite{Zhou-IJCAI-2018}. 

\begin{figure}[t!]
    \centering
    \includegraphics[width=0.48\textwidth]{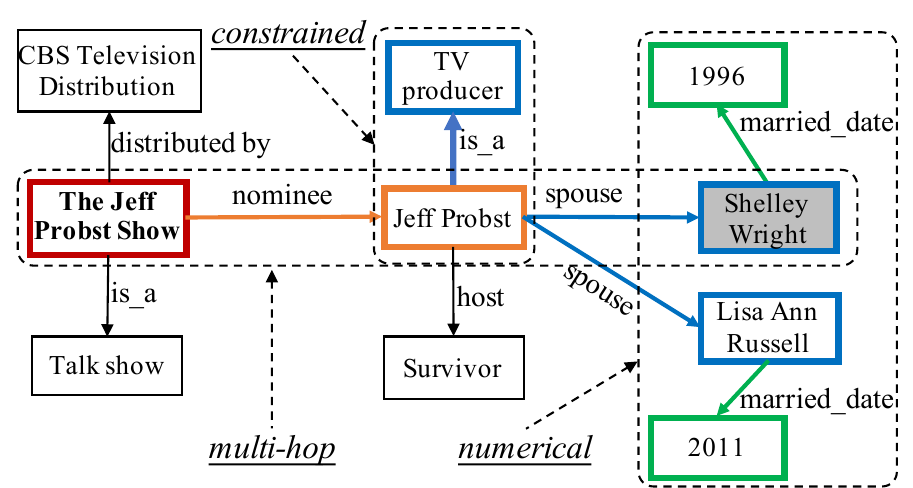}
    \caption{An example of complex KBQA for the question ``\textit{Who is the first wife of TV producer that was nominated for The Jeff Probst Show?}''. 
    We present the related KB subgraph for this question.
    The ground truth path heading to the answer is annotated with colored borders. 
    The topic entity and the answer entity are shown in the bold font and shaded box respectively.
    ``multi-hop'' reasoning, ``constrained'' relations, and ``numerical'' operation are highlighted in black dotted box. We use different colors to indicate different reasoning hops to reach each entity from the topic entity.
    }
    \label{fig:example}
\end{figure}

Early work on KBQA focused on answering a simple question, where only a single fact is involved. 
For example, ``\textit{Who was the nominee of The Jeff Probst Show?}'' is a simple question which includes the subject ``\textit{The Jeff Probst Show}'', the relation ``\textit{nominee}'' and queries about the object entity ``\textit{Jeff Probst}'' of fact ``\textit{(The Jeff Probst Show, nominee, Jeff Probst)}'' in KBs. 
It is not trivial to retrieve the correct entity from the large-scale KBs, which consists of millions or even billions of facts.
Therefore, researchers have spent much effort in proposing different models to answer simple questions over KBs~\cite{Bordes-ArXiv-2015,Dong-ACL-2015,Yu-ACL-2017,Petrochuk-EMNLP-2018,Hu-TKDE-2018}.

 

\begin{figure*}[t!]
    \centering
    \includegraphics[scale=0.42]{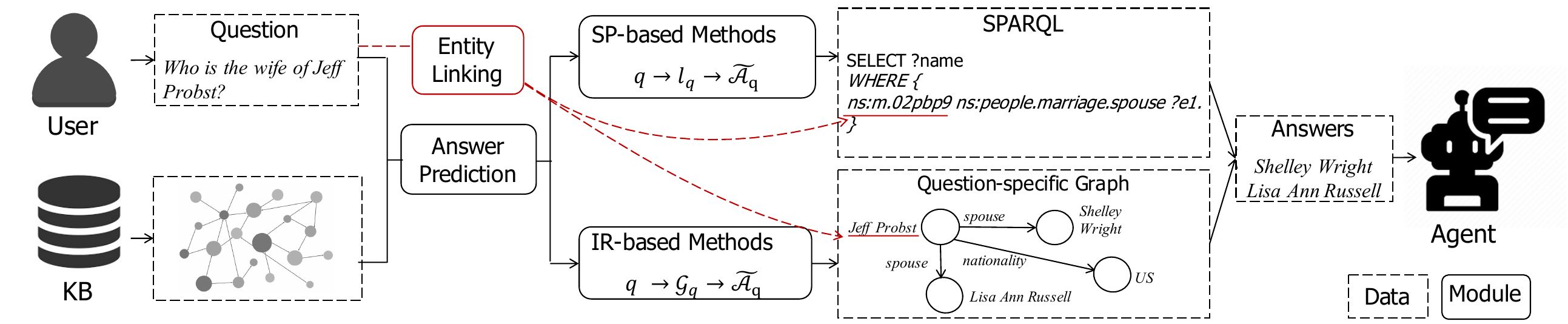}
    \caption{Architecture of KBQA systems. The entity linking procedure is shown in red color.
    }
    \label{fig:pipeline}
\end{figure*}

Recently, researchers started paying  more attention to answering \textbf{\emph{complex questions}} over KBs, \ie the complex KBQA task~\cite{Hu-EMNLP-18,Luo-EMNLP-2018}.
Complex questions usually contain multiple subjects, express compound relations, or include numerical operations. 
Take the question in Figure~\ref{fig:example} as an example.
This example question starts with the subject ``\textit{The Jeff Probst Show}''.
Instead of querying a single fact, the question requires the composition of two relations, namely, ``\textit{nominee}'' and ``\textit{spouse}''.
This query is also associated with an entity type constraint ``\textit{(Jeff Probst, is a, TV producer)}''. 
The final answer should be further aggregated by selecting the possible candidates with the earliest marriage date.
Generally, complex questions are questions involving \textbf{\emph{multi-hop reasoning}},  \textbf{\emph{constrained relations}} or \textbf{\emph{numerical operations}}.

Tracing back to the solutions for simple KBQA task, a number of studies from two mainstream approaches have been proposed. 
We show the overall architecture of simple KBQA systems in Figure~\ref{fig:pipeline}.
These two approaches first recognize the subject in a question and link it to an entity in the KB (referred to as the \textbf{\emph{topic entity}}).
Then they derive the answers within the neighborhood of the topic entity by either executing a parsed logic form or reasoning in a question-specific graph extracted from the KB.
The two categories of methods are commonly known as \textbf{\emph{semantic parsing-based}~(SP-based)} methods and \textbf{\emph{information  retrieval-based}~(IR-based)} methods in prior work~\cite{Bordes-ArXiv-2015,Dong-ACL-2015,Hu-TKDE-2018,GrailQA-2020}.
They design different working mechanisms to solve the KBQA task. 
The former approach represents a question by a symbolic logic form, and then executes it against the KB to obtain the final answers. 
The latter approach constructs a question-specific graph delivering the comprehensive information related to the question, and generates the final answers based on the extracted graph.

However, when applying the two mainstream approaches to the complex KBQA task, complex questions bring in challenges on different parts of the approaches:

\begin{itemize}
    \item Parsers used in existing SP-based methods are difficult to cover diverse complex queries (\eg multi-hop reasoning, constrained relations). 
    Similarly, previous IR-based methods may fail to answer a complex query, as the answer is generated without traceable reasoning.
    \item More relations and subjects in complex questions indicate a larger search space of potential logic forms for parsing, which will dramatically increase the computational cost.
    Meanwhile, more relations and subjects prevent IR-based methods from retrieving all relevant facts for reasoning, which makes the common incomplete KB issue become severer.
    \item 
    When questions become complicated from both semantic and syntactic aspects, models are required to have strong capabilities of natural language understanding and generalization.
    Comparing the question ``\textit{Who is the first wife of TV producer that was nominated for The Jeff Probst Show?}'' with another question ``\textit{Who is the wife of the first TV producer that was nominated for The Jeff Probst Show?}'', the models are supposed to understand that the ordinal number ``\textit{first}'' is used to constrain ``\textit{wife}'' or the phrase ``\textit{TV producer}''. 
    \item 
    Generally, only question-answer pairs are provided.
    This indicates SP-based methods and IR-based methods have to be trained without the annotation of correct logic forms and reasoning paths.
    Such weak supervision signals bring difficulties to both approaches due to the lack of guidance in intermediate reasoning process.
\end{itemize}

Regarding the related surveys, we observe Wu~\etal~\cite{Wu-CCK-2019} and Chakraborty~\etal~\cite{Chakraborty-arXiv-2019} reviewed the existing work on simple KBQA. 
Gu~\etal~\cite{gu:akbc2022} provided a semantic parsing perspective to KBQA task.
Furthermore, Fu~\etal~\cite{Fu-Arxiv-2020} investigated the current advances on complex KBQA.
They provided a general view of advanced methods only from the perspective of techniques and more focused on application scenarios in e-commerce domain.
Different from these surveys, our work tries to identify the challenges encountered in previous studies, and extensively discuss existing solutions in a comprehensive and well-organized manner. 
It is worth noting that this survey is an extended version of the short survey~\cite{Lan-IJCAI-2021}. 
As a comparison, this survey has several main differences: 
(1) We add more recent-published papers and refine the description of challenges as well as solutions with a fine-grained taxonomy.
(2) We provide deep discussions of the two mainstream categories including a comprehensive comparison of their core modules and a unified paradigm of neural symbolic reasoning.
(3) We add a new section to discuss the role of cutting-edge pre-trained language techniques for complex KBQA and give a more thorough outlook on several promising research directions.
(4) We attach a companion page with all methods, chronological information, open resources and reported results to help readers quickly catch up with the development of this task.
(5) We introduce traditional approaches, preliminary knowledge, and evaluation protocol of KBQA in multiple aspects and a more concrete way, which goes far beyond the scope of the short survey.




\ignore{This paper is organized as follows.
First, we introduce multiple available datasets for conducting complex KBQA researches.
Second, we summarize and compare the advanced approaches.
Finally, we conclude and discuss future research directions.}

The remainder of this survey is organized as follows.
We first introduce preliminary knowledge in Section~\ref{sec:task_dataset}.
Next, we describe the two mainstream categories of methods for complex KBQA in Section~\ref{sec-methods}. 
Following the categorization, we figure out typical challenges and solutions for SP-based and IR-based methods in Section~\ref{sec-SP} and Section~\ref{sec-IR}, respectively. 
We highlight the impact of pre-trianed language models on complex KBQA in Section~\ref{sec:PLM}. 
We summarize datasets, and relevant resource in Section~\ref{sec:evaluation_resource}. 
Finally, we discuss recent trends and conclude in Section~\ref{sec:direction} and Section~\ref{sec:con}.

%% file: sec-back.tex
\section{Preliminary}
\label{sec:task_dataset}
In this section, we first briefly introduce KBs and task formulation of KBQA, then we talk about the traditional approaches for KBQA systems.

\ignore{
Formally speaking, a KB consists of a set of facts, where the subject and object are connected by their relation.
All the subjects and objects in facts form the entity set of the KB.
A complex question is a natural language question that is in the format of a sequence of tokens.
We define the complex KBQA task as follows.
Given a question and an associated KB, the system is supposed to retrieve the entities or aggregate the retrieved entities from the entity set as the corresponding answers.
Generally, the ground truth answers of~\jhcomment{to} the questions are provided to train a complex KBQA system.
}

\subsection{Knowledge Base}

As mentioned earlier, KB is usually in the format of triples. They are designed to support modeling  relationships between entities. 
Take Freebase~\cite{freebase} as an example for KB.
Each entity in Freebase has a unique ID (refereed to as \textbf{\textit{mid}}), 
one or more types, and uses properties from these types in order to provide facts~\cite{Thomas-WWW-2016}. 
For example, the Freebase entity for person Jeff Probst has the mid ``$m.02pbp9$''\footnote{PREFIX: $<http://rdf.freebase.com/ns/>$} and the type ``$people.marriage$'' that allows the entity to have a fact with ``$people.marriage.spouse$'' as the property and ``$m.0j6d0bg$'' (psychotherapist Shelley Wright) as the value.
Freebase incorporates compound value types (CVTs) to represent n-ary ($n > 2$) relational facts~\cite{Bollacker-SIGMOD-2008} like ``\textit{Jeff Probst was married to Shelley Wright in 1996}'', where three entities, namely ``\textit{Jeff Probst}'', ``\textit{Shelley Wright}'', and ``\textit{1996}'', are involved in a single statement.
Different from entities which can be aligned with real world objects or concepts, CVTs are artificially created for such n-ary facts.

In practice, large-scale open KBs (\eg Freebase and DBPedia) are published under Resource Description Framework (RDF) to support structured query language~\cite{Cyganiak-WWW-2014,Thomas-WWW-2016}.
To facilitate access to large-scale KBs, the query language SPARQL is frequently used to retrieve and manipulate data stored in KBs~\cite{Thomas-WWW-2016}.
In Figure~\ref{fig:pipeline}, we have shown an executable SPARQL to obtain the spouses of entity ``\textit{Jeff Probst}''.

Different KBs are designed with different purposes, and have varying properties under different schema design.
For example, Freebase is created mainly by community members and harvested from many resources including Wikipedia.
YAGO~\cite{yago-2016} takes  Wikipedia and WordNet~\cite{wordnet-2015} as the knowledge resources and covers taxonomy of more general concepts.
WikiData~\cite{Thomas-WWW-2016} is a multilingual KB which integrates multiple resources of KBs with high coverage and quality. 
A more comprehensive comparison between open KBs is available at~\cite{farber-2015}. 

\subsection{Task Formulation}
\ignore{
Formally, a KB consisting of a set of facts is given as input, where the subject and object are connected by their relation. 
We denote it as $\mathcal{G} = \{\langle e, r, e' \rangle | e, e' \in \mathcal{E}, r \in \mathcal{R}\}$, where $\langle e, r, e' \rangle$ denotes that relation $r$ exists between subject $e$ and object $e'$, $\mathcal{E}$ and $\mathcal{R}$ denote the entity set and relation set, respectively.~\glcomment{I think we'd better move this part to KB definition.}
}
Formally, we denote a KB as $\mathcal{G} = \{\langle e, r, e' \rangle | e, e' \in \mathcal{E}, r \in \mathcal{R}\}$, where $\langle e, r, e' \rangle$ denotes that relation $r$ exists between subject $e$ and object $e'$, $\mathcal{E}$ and $\mathcal{R}$ denote the entity set and relation set, respectively.

Given the available KB $\mathcal{G}$, this task aims to answer  natural language questions $q = \{w_1, w_2, ..., w_m\}$ in the format of a sequence of tokens (typically organized with a unique vocabulary $\mathcal{V}$) and we denote the predicted answers as $\Tilde{\mathcal{A}}_q$. 
Specially, existing studies assume the correct answers $\mathcal{A}_q$ can be derived from the entity set $\mathcal{E}$ of the KB or a natural language sequence (\ie the surface name of entities).
Unlike answers of simple KBQA task, which are entities directly connected to the topic entity, the answers of the complex KBQA task are entities multiple hops away from the topic entities or even the aggregation of them.
Generally, a KBQA system is trained using a dataset $\mathcal{D} = \{(q, \mathcal{A}_{q})\}$.



\subsection{Traditional Approaches}
General KBQA systems for simple questions have a pipeline framework as displayed in Figure~\ref{fig:pipeline}.
The preliminary step is to identify topic entity $e_q$ of the question $q$, which aims at linking a question to its related entities in the KBs.
In this step, named entity recognition, disambiguation and linking are performed.
It is usually done using some off-the-shelf entity linking tools, such as S-MART~\cite{Yang-ACL-2015}, DBpedia Spotlight~\cite{Daiber-IS-2013}, and AIDA~\cite{Yosef-VLDB-2011}.
Subsequently, an answer prediction module is leveraged to predict the answer $\Tilde{\mathcal{A}}_q$ taking $q$ as the input.

For simple KBQA task, the predicted answers are usually located within the neighborhood of the topic entities.
Different features, as well as methods, are proposed to rank these candidate entities.
Early attempts on solving simple KBQA task employed existing semantic parsing tools to parse a simple natural language question into an uninstantiated logic form, and then adapted it to KB schema by aligning the lexicons.
This step results in an executable logic form $l_q$ for $q$. 
In detail, the existing semantic parsing tools usually follow Combinatory Categorical Grammars (CCGs)~\cite{Cai-ACL-2013, Kwiatkowski-EMNLP-2013,Reddy-TACL-2014} to build domain-independent logic forms.
Then different methods~\cite{Unger-WWW-2012,Cai-ACL-2013,Kwiatkowski-EMNLP-2013,Berant-EMNLP-2013,Reddy-TACL-2014} are proposed to perform schema matching and lexicon extension, which results in logic forms grounded with KB schema.
For simple KBQA task, this logic form is usually a single triple starting from the topic entity and connecting to the answer entities.
As early methods heavily rely on rule-based mapping, which is hard to be generalized to large-scale datasets~\cite{Yih-ACL-2014,Yao-ACL-2014,Yih-ACL-2016}.
Follow-up work proposed some scoring functions to automatically learn the lexicon coverage between the logic forms and the questions~\cite{Yih-ACL-2015,Reddy-TACL-2016}.
With the development of deep learning, several advanced neural networks such as Convolutional Neural Network~\cite{Yin-COLING-2016},  Hierarchical Residual BiLSTM~\cite{Yu-ACL-2017}, Match-Aggregation Module~\cite{Lan-TASLP-2019}, and Neural Module Network~\cite{Andreas-NAACL-2016} are utilized to measure the semantic similarities.
This line of work is known as semantic parsing-based methods.

Information retrieval-based methods were also developed over the decades. 
They retrieve a question-specific graph $\mathcal{G}_{q}$ from the entire KB. 
Generally, entities one hop away from the topic entity and their connected relations form the subgraph for solving a simple question.
The question and candidate answers in the subgraph can be represented as low-dimensional dense vectors.
Different ranking functions are proposed to rank these candidate answers and top-ranked entities are considered as the predicted answers~\cite{Bordes-PKDD-2014,Bordes-EMNLP-2014,Bordes-ArXiv-2015, Petrochuk-EMNLP-2018}.
Afterwards, Memory Network~\cite{Sukhbaatar-NIPS-2015} is employed to generate the final answer entities~\cite{KVMem-EMNLP-2016,Jain-NAACL-2016}.
More recent work~\cite{Dong-ACL-2015,Hao-ACL-2017,Chen-NAACL-2019} employs attention mechanism or multi-column modules to this framework to boost the ranking accuracy.
In Figure~\ref{fig:pipeline}, we have displayed different pipelines and intermediate outputs of the two methods. 

%% file: sec-pre.tex
\begin{figure}[!t]
 \centering
 \includegraphics[width=0.5\textwidth]{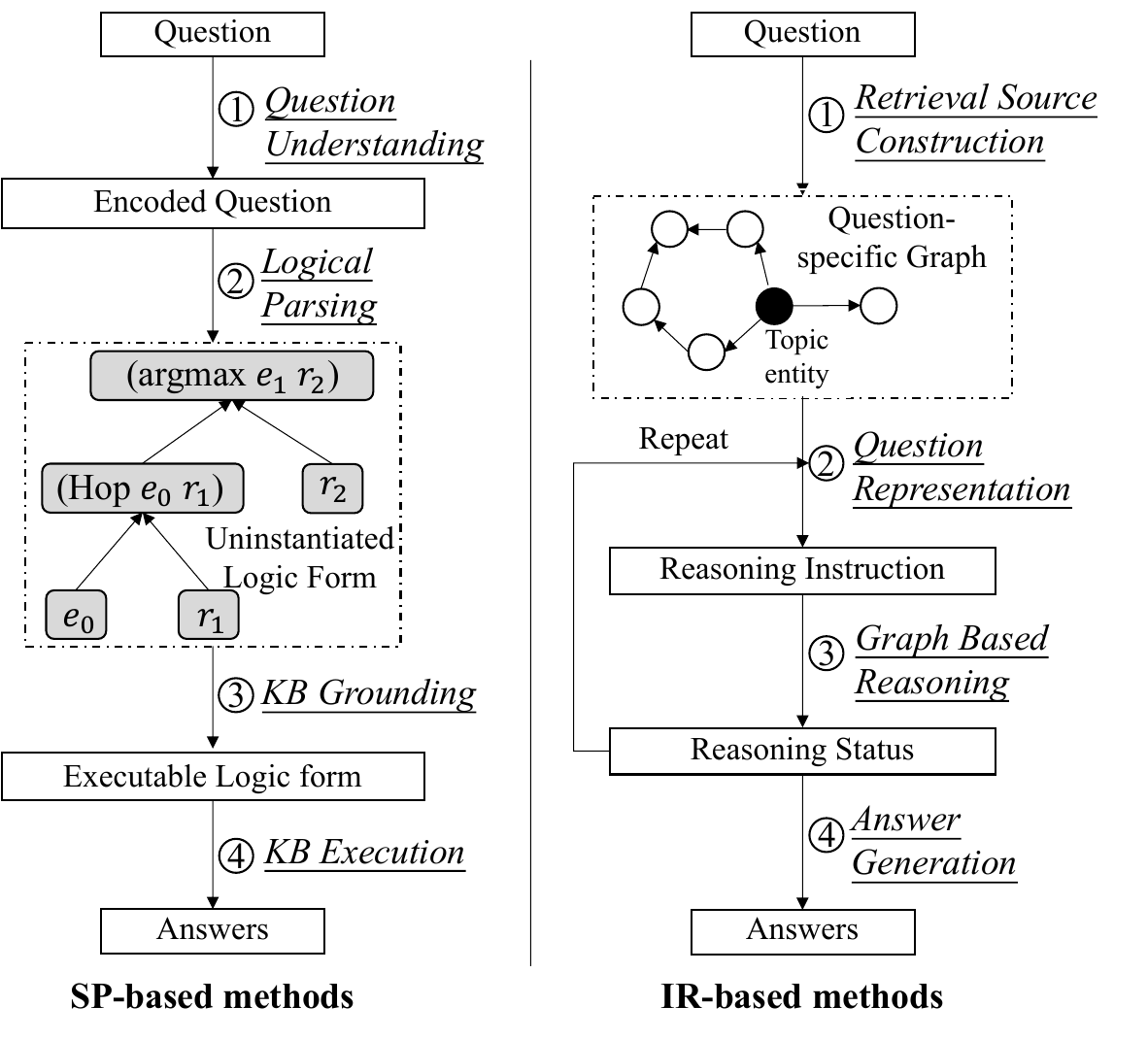}
 \centering
 \caption{Illustration of two mainstream approaches for complex KBQA.
 }
 \label{fig-methods}
\end{figure}

\section {Two Mainstream Approaches}
\label{sec-methods}

Complex KBQA systems follow the same overall architecture as shown in Figure~\ref{fig:pipeline}, where the entity linking is first performed. 
Subsequently, as introduced in Section~\ref{sec:intro}, SP-based and IR-based methods are two mainstream approaches to answering complex questions. 
SP-based methods parse a question into a logic form and execute it against KBs for finding the answers.
IR-based methods retrieve a question-specific graph and apply some ranking algorithms to select entities from top positions or directly generate answers with text decoder.
To summarize, the two approaches follow either a \emph{parse-then-execute} paradigm or 
a \emph{retrieve-and-generate} paradigm.
To show the difference between the two paradigms, we illustrate their question answering procedures with detailed modules in Figure~\ref{fig-methods}.

\subsection{Semantic Parsing-based Methods} 
As shown in Figure~\ref{fig-methods}, we summarize the procedure of SP-based methods into the following four modules:
\begin{enumerate}[label=(\arabic*)]
    \item They understand a question via a \underline{\emph{question}} \underline{\emph{understanding}} module, which is to conduct the semantic and syntactic analysis and obtain an encoded question for the subsequent parsing step.
    We denote this module as follows:
    \begin{align*}
        \Tilde{q} = \textit{Question\_Understanding}(q),
    \end{align*}
    where $\Tilde{q}$ is the encoded question that captures semantic and syntactic information of the natural language question.
    It can be distributed representation, structural representation or their combination.
    Intuitively, neural networks (\eg  LSTM~\cite{Hochreiter-NeuralComput-1997}, GRU~\cite{Cho-SSST-2014}, and PLMs) are employed to act as the question understanding module and obtain hidden states to represent the question. 
    Meanwhile, some syntactic parsing is performed to extract structural properties of the question.
    \item A \underline{\emph{logical parsing}} module is utilized to transfer the encoded question into an uninstantiated logic form:
    \begin{align*}
        \bar{l}_q = \textit{Logical\_Parsing}(\Tilde{q}),
    \end{align*}
    where $\bar{l}_q$ is the uninstantiated logic form without the detailed entities and relations filled in.
    The grammar and constituents of logic forms can be different with specific designs of a system.
    Here, $\bar{l}_q$ can be obtained by either generating a sequence of tokens or ranking a set of candidates.
    In practice, Seq2seq models or featured based ranking models are employed to generate $\bar{l}_q$ based on the encoded question. 
    \item To execute against KBs, the logic form is further instantiated and validated by conducting some semantic alignments to structured KBs via \underline{\emph{KB grounding}}. 
    Note that, in some work~\cite{Yih-ACL-2015,Liang-ACL-17}, the logical parsing and KB grounding are simultaneously performed, where logic forms are validated in KBs while partially parsed:
    \begin{align*}
        l_q = \textit{KB\_Grounding}(\bar{l}_q, \mathcal{G}).
    \end{align*}
    After this step, $\bar{l}_q$ is instantiated with the entities and relations in $\mathcal{G}$ so that we obtain an executable logic form $l_q$. 
    It is worth noting that $l_q$ always contains $e_q$, which are detected via entity linking module.
    Its format is not restricted to the SPARQL query but always transferable to SPARQL. 
    \item Eventually, the parsed logic form is executed against KBs to generate predicted answers via a \underline{\emph{KB execution}} module:
    \begin{align*}
        \Tilde{\mathcal{A}}_q = \textit{KB\_Execution}(l_q),
    \end{align*}
    where $\Tilde{\mathcal{A}}_q$ is the predicted answers for the given question $q$.
    This module is usually implemented via an existing executor.
\end{enumerate}

During training, the logic form $l_q$ is treated as the intermediate output.
The methods are trained using the KBQA datasets in the format of $\mathcal{D} = \{(q, \mathcal{A}_q)\}$, where the objective is set to generate a logic form matching the semantics of the question and resulting in correct answers.


\subsection{Information Retrieval-based Methods} 
Similarly, we summarize the procedure of IR-based methods into four modules as illustrated in Figure~\ref{fig-methods}:
\begin{enumerate}[label=(\arabic*)]
    \item Starting from the topic entity $e_q$, the system first extracts a question-specific graph from KBs. 
    Ideally, this graph includes all question-related entities and relations as nodes and edges respectively. 
    Without explicitly generating an executable logic form, IR-based methods perform reasoning over the graph. 
    We represent a \underline{\emph{retrieval source construction}} module taking as input of both the question and KB as:
    \begin{align*}
        \mathcal{G}_{q} = \textit{Retrieval\_Source\_Construction}(q, \mathcal{G}),
    \end{align*}
    where $\mathcal{G}_{q}$ is the question-specific graph extracted from $\mathcal{G}$. As the size of subgraph grow exponentially with the distance to topic entities, some filtering tricks (\eg personalized Pagerank) are adopted to keep the graph size in a computation-affordable scale~\cite{Sun-EMNLP-2018,Xiong-ACL-2019}.
    \item Next, the system encodes input questions via a \underline{\emph{question representation}} module.
    This module analyzes the semantics of the question and outputs reasoning instructions, which are usually represented as vectors. Typically, question $q$ are encoded into hidden vectors $\bm{q}$ with neural networks (\eg LSTM, GRU, and PLMs) and then combined with other methods (\eg attention mechanism) to generate a vector as instruction:
    \begin{align*}
        \bm{i}^{(k)} = \textit{Question\_Representation}(\bm{i}^{(k-1)}, q, \mathcal{G}_q)
    \end{align*}
    Here, $\{\bm{i}^{(k)}, k=1,..., n\}$ is the instruction vector of $k$-th reasoning that encodes the semantic and syntactic information of the natural language question. 
    Both multi-step reasoning and one-step matching are applicable, which results in varying reasoning steps $n$.
    \item A \underline{\emph{graph based reasoning}} module conducts semantic matching 
    via vector based computation to propagate and aggregate the information along the neighboring entities within the graph.
    The reasoning status $\{s^{(k)}, k=1, ..., n\}$, which has diverse definitions in different methods (\eg distributions of predicted entities and representations of relations), is updated based on the reasoning instruction:
    \begin{align*}
        s^{(k)} = \textit{Graph\_Based\_Reasoning}(s^{(k-1)}, \bm{i}^{(k)}, \mathcal{G}_{q}),
    \end{align*}
    where $s^{(k)}$ is the reasoning status which is considered as the status of $k$-th reasoning step on graph. 
    Recently, several studies~\cite{Jain-NAACL-2016,Chen-NAACL-2019} repeat Step (2) and (3) for multiple times to perform the reasoning.
    \item An \underline{\emph{answer generation}}  module is utilized to generate answers according to the reasoning status at the end of reasoning. There are mainly two types of such generators: (1) entity ranking generator which ranks the entities to obtain top-ranked entities as predicted answers, (2) text generator which generates free text answers with vocabulary $\mathcal{V}$. This module can be formalized as:
    \begin{align*}
        \Tilde{\mathcal{A}}_q = \textit{Answer\_Generation}(s^{(n)}, \mathcal{G}_q, \mathcal{V}),
    \end{align*}
    where $s^{(n)}$ denotes the reasoning status at the last step.
    
    In the entity ranking paradigm, the entities contained in $\mathcal{G}_q$ are candidates for answer prediction $\Tilde{\mathcal{A}}_q$. In many cases, $\Tilde{\mathcal{A}}_q$ is obtained through selecting the entities with a score larger than the pre-defined threshold, where the score is derived from $s^{(n)}$. While in text generation paradigm, the answers are generated from vocabulary $\mathcal{V}$ as a sequence of tokens.
\end{enumerate}

During training, the objective of entity ranking generator is usually to rank the correct entities higher than others in $\mathcal{G}_q$. In comparison, the text generator is usually trained to generate gold answers (name of correct entities).
\subsection{A Comparison of Core Modules}

Comparing the procedures of SP-based and IR-based methods, we note that these two methods have different designs of core modules and working mechanisms, but they also share similarities from multiple aspects.

\paratitle{Difference}: SP-based methods rely heavily on the logical parsing module, which produces an expressive logic form for each question. 
In practice, many commercial KBQA systems developed upon SP-based methods require expertise to provide feedback to the generated logic forms so that the system can be further improved~\cite{Diefenbach-KIS-2018}.
However, considering the expensive cost and expertise for obtaining annotated logic form, SP-based methods are usually trained under a weak supervision setting in research. 
Compared with IR-based methods, SP-based methods have the advantage of showing interpretability with explicit evidence about reasoning and defending perturbation of the question. 
However, the logical parsing module is bound by the design of the logic form and the capability of the parsing techniques, which is the key of performance improvement. 

As a comparison, IR-based methods first employ a retrieval module to obtain the question-specific graph, and then conduct complex reasoning on the graph structure with graph based reasoning module. 
The answers are eventually predicted via an answer generation module.
The performance of the IR-based methods partially depends on the recall of the retrieval module as the subsequent reasoning takes the retrieved graph as input. 
Meanwhile, the graph based reasoning and answer generation module play key roles in making accurate predictions.
Instead of generating logic forms, IR-based methods directly generate entities or free text as predictions. So they naturally fit into the end-to-end training paradigm and could be optimized easier compared with SP-based methods.
Nevertheless, the blackbox style of the reasoning module makes the reasoning process less interpretable, which decreases the robustness and hinders users from interacting with the system.

\begin{figure*}
	\footnotesize
	\begin{forest}
		for tree={
			forked edges,
			grow'=0,
			draw,
			rounded corners,
			node options={align=center},
			calign=edge midpoint,
		},
	    [SP-based method, text width=1.5cm, for tree={fill=white!20}
			[Question understanding, text width=1.8cm, for tree={fill=red!20}
				[Understanding complex semantics of questions, text width=2.4cm, for tree={fill=red!20}
				    [
				    Incorporating structure property of questions to seq2seq generation: parse with dependency paths~\cite{Luo-EMNLP-2018,Abujabal-WWW-17,Abujabal-WWW-2018}; parse with AMR~\cite{Kapanipathi-AAAI-2021} and skeleton-based parsing~\cite{Skeleton-AAAI-2020},
				    text width=10.5cm, node options={align=left}
				    ]
				]
				[Understanding complex syntax of queries, text width=2.4cm, for tree={fill=red!20}
					[
					Incorporating structure property of logic forms to feature-based ranking: match with structure properties pf queries~\cite{Zhu-Neuro-2020}; Tree-LSTM~\cite{Zafar-ESWC-2018}; predict structure of queries as intermediate state~\cite{Chen-IJCAI-2020},
					text width=10.5cm, node options={align=left}
					]
				]
			]
			[Logical parsing, text width=1.8cm, for tree={fill=orange!20}
				[Parsing complex queries, text width=2.4cm, for tree={fill=orange!20}
					[
					Designing logic forms via pre-defined query templates: three query templates~\cite{Bast-CIKM-15}; temporal query templates~\cite{Jia-CIKM-2018},
					text width=10.5cm, node options={align=left}
					]
					[
					Designing logic forms with flexible combining rules: query graph~\cite{Yih-ACL-2015}; query graphs with more aggregation operators~\cite{Abujabal-WWW-17,Hu-TKDE-2018},
					text width=10.5cm, node options={align=left}
					]
				]
			]
			[KB grounding, text width=1.8cm, for tree={fill=cyan!20}
				[Grounding with large search space, text width=2.4cm, for tree={fill=cyan!20}
					[
					Decomposing a complex question to sub-questions: template decomposition~\cite{Zheng-VLDB-2018};  Bhutani\etal~\cite{Bhutani-CIKM-2019},
					text width=10.5cm, node options={align=left}
					]
					[
					Expanding a logic form by iteration: unrestricted-hop relation extraction~\cite{Chen-NAACL-2019}; query graph generator~\cite{Lan-ICDM-2019,Lan-ACL-2020},
					text width=10.5cm, node options={align=left}
					]
				]
			]
			[Training procedure, text width=1.8cm, for tree={fill=lime!20}
				[Training with sparse reward, text width=2.4cm, for tree={fill=lime!20}
					[
					Augmenting final reward with enriched features: answer types~\cite{Saha-TACL-2019},
					text width=10.5cm, node options={align=left}
					]
					[
					Augmenting intermediate reward with enriched critics: option-based hierarchical framework~\cite{Qiu-CIKM-2020} and critics with hand-crafted rules~\cite{Qiu-WSDM-2020},
					text width=10.5cm, node options={align=left}
					]
				]
				[Training with spurious reasoning, text width=2.4cm, for tree={fill=lime!20}
					[
					Stabilizing training processing with high-reward logic forms: bootstrap training~\cite{Liang-ACL-17}; replay high-reward logic forms in memory buff~\cite{Hua-JWS-2020},
					text width=10.5cm, node options={align=left}
					]
				]
			]
		]
	\end{forest}
	\caption{The main content of SP-based methods. The hierarchical structure is arranged with: SP-based method $\rightarrow$ module $\rightarrow$ challenge  $\rightarrow$ solution.}
    \label{fig:taxonomy-sp}
\end{figure*}
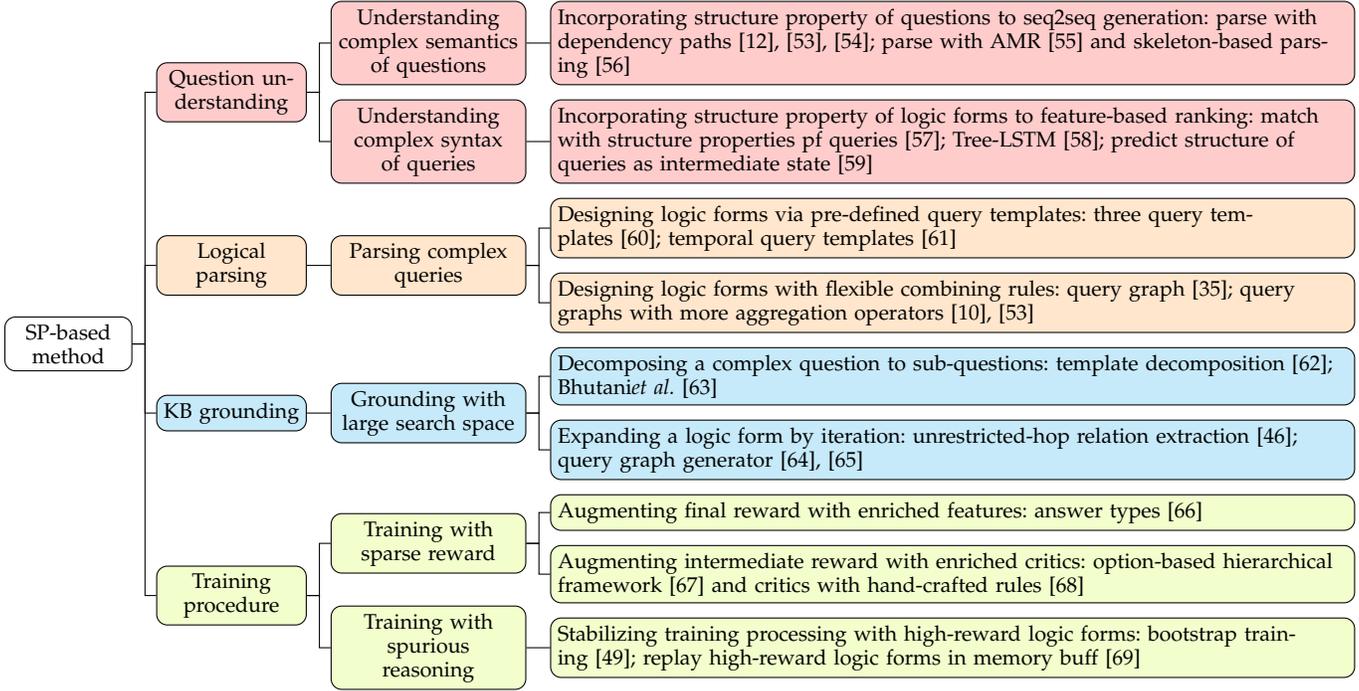

\paratitle{Similarity}: SP-based and IR-based methods both contain parameter-free modules, which are KB grounding, KB execution modules, and retrieval source construction module, respectively.
While they are generally not learned from KBQA datasets, their performance has a great impact on the final KBQA performance.
Both categories of methods make use of detected topic entities. SP-based methods leverage them to instantiate the logic form in the KB grounding module, while IR-based methods utilize them to narrow down the reasoning scale in the retrieval source module.
Besides, both SP-based and IR-based methods emphasize the importance of natural language understanding with a question understanding (representation) module. 
The output of such modules substantially influences the subsequent parsing or reasoning process.


\subsection{A Unified Paradigm - Neural Symbolic Reasoning}
\label{sec-NSR}

\ignore{
\begin{figure}[!t]
 \centering
 \includegraphics[width=0.5\textwidth]{Figures/nsm.pdf}
 \centering
 \caption{Integration of neural symbolic machine for knowledge base question answering.
 }
 \label{fig:nsm}
\end{figure}
}
In recent years, neural symbolic reasoning~(NSR) has become a hot topic in machine learning. 
\emph{It describes a type of hybrid systems that apply the high-efficiency of connectionism (the neural system) and the generalization of symbolism (the symbolic system) to integrate learning and reasoning effectively}~\cite{gal:vldb2015,lamb:ijcai2020,Zhang-arxiv-2021}.
Relevant techniques are widely applied to intelligent applications like question answering~\cite{gal:www2013}, semantic parsing~\cite{sun:aaai2020}. As for complex KBQA, NSR techniques are helpful in addressing some challenges for SP-based and IR-based methods. Furthermore, NSR may be a potential paradigm to unify both SP-based and IR-based methods.

The symbolic system and neural system play different roles in KBQA tasks.
The symbolic system usually takes the KBs and grammar rules as inputs, searches for the solution space for question answering, and performs reasoning for the results.
In comparison, the neural system takes the natural language questions and elements in KB as inputs, learns a neural network model for a specific task, and serves the reasoning in latent space.
In this way, NSR-based methods could reason with powerful neural networks in latent space, meanwhile providing explicit inference evidence to explain the results as well as the reasoning process.
For SP-based methods, the logical parsing and KB grounding modules act as the symbolic system to interact with KBs and search for potential logical forms.
The question understanding module usually acts as the neural system to learn the semantic matching between the given question and the potential logical forms~\cite{Yih-ACL-2016,Liang-ACL-17,Luo-EMNLP-2018}.
For IR-based methods, the retrieval source construction and graph based reasoning modules link to the symbolic system, while the neural system usually consists of question representation and graph based reasoning modules~\cite{Sun-EMNLP-2018,KVMem-EMNLP-2016,Xiong-ACL-2019}.

The benefits of applying neural symbolic reasoning to complex KBQA come from the following aspects:
\textbf{1) Symbolic system facilitates discrete inferences on the structured data.} Symbolic system helps narrow down the search space for complex questions, increase the interpretability of reasoning process~\cite{Bao-COLING-2016} and compositional generalization capability of systems~\cite{GrailQA-2020}.
For example, Bao et al.~\cite{Bao-COLING-2016} and Lan et al.~\cite{Lan-ACL-2020} developed symbolic systems to cope with a large range of questions and integrate diverse modalities.
\textbf{2) Neural system facilitates modeling heterogeneous and imperfect data.}
Neural system deals with the diversity of natural language expressions of complex questions~\cite{Sun-EMNLP-2018}, manipulates with heterogeneous data~\cite{Ye-ACL-22} (\eg complex questions, entities, relations, and even generated templates) and even infers relations that are missing from the incomplete KB~\cite{Apoorv-ACL-2020}.
It has been proven to be effective in solving the above issues in both SP-based and IR-based methods~\cite{Yih-ACL-2016,Sun-EMNLP-2018}. 
\textbf{3) Neural symbolic reasoning (NSR) can take advantage of both neural systems and symbolic systems.} 
For SP-based methods, Liang et al.~\cite{Liang-ACL-17} designed the subset of $\lambda$-calculus as the search space of logical forms in the symbolic system to fit the nature of complex queries in complex KBQA, which further constrains the generation of programs through seq2seq neural models.
Similarly, in IR-based methods, Sun et al.~\cite{Sun-EMNLP-2019} developed PullNet, a neural symbolic machine that conducts (symbolic) reasoning graph expanding along with the graph-based (neural) reasoning process, and achieved promising performance.

As we can see, complex KBQA systems show a trend to connect with and benefit from neural symbolic reasoning. The two mainstream approaches can be unified with the neural symbolic reasoning paradigm, which differs in the detailed designs of the symbolic system and neural system.
More discussion can be found in recent study~\cite{Zhang-arxiv-2021}.

%% file: sec-main.tex
\label{sec:challenge_solution}

\input{sec-sp}

\input{sec-ir}

%% file: sec-sp.tex
\section{Semantic Parsing-based Methods}
\label{sec-SP}
In this part, we discuss the challenges and solutions for semantic parsing-based methods. The taxonomy of challenges and solutions can be visualized with Figure~\ref{fig:taxonomy-sp}.


\subsection{Overview}
As introduced in Section~\ref{sec-methods}, SP-based methods follow a parse-then-execute procedure via a series of modules, namely question understanding, logical parsing, KB grounding, and KB execution. These modules encounter different challenges for complex KBQA.
Firstly, question understanding becomes more difficult when the questions are complicated in both semantic and syntactic aspects.
Secondly, logical parsing has to cover diverse query types of complex questions. 
Moreover, a complex question involving more relations and subjects will dramatically increase the possible search space for parsing.
Thirdly, the manual annotation of logic forms is expensive and labor-intensive, and it is challenging to train the SP-based methods with weak supervision signals (\ie question-answer pairs). 

In the following parts, we will introduce how prior studies deal with these challenges and summarize advanced techniques proposed by them.

\begin{figure}[htbp]
 \centering
 \includegraphics[width=0.48\textwidth]{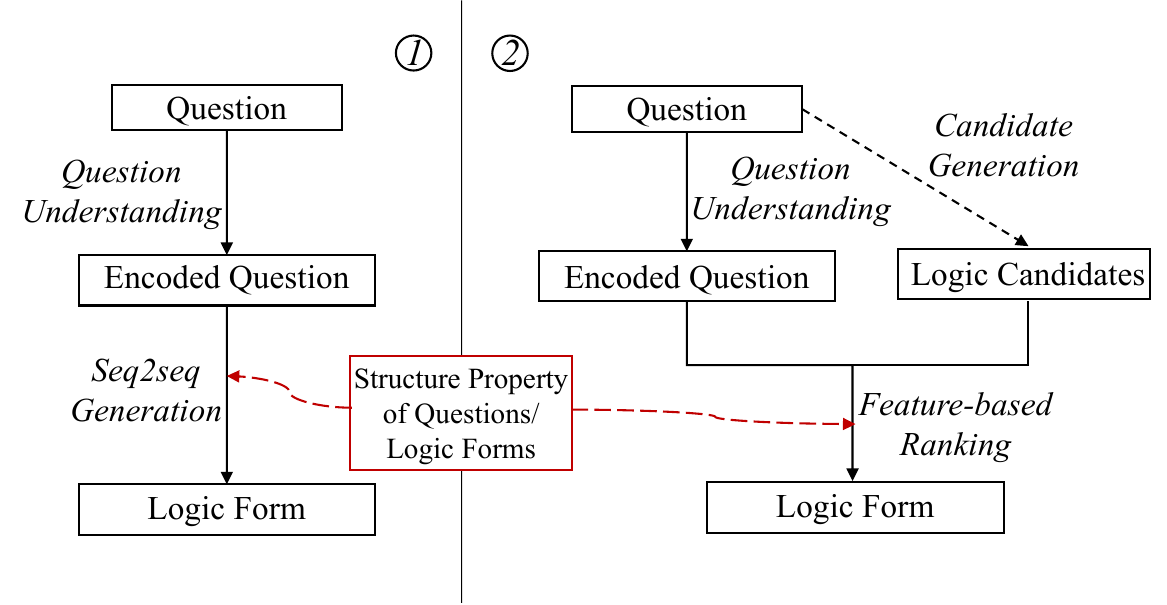}
 \centering
 \caption{Illustration of two lines of research which leverage structure properties for better understanding of complex question.
 }
 \label{fig-syntactic}
\end{figure}

\subsection{Understanding Complex Semantics and Syntax} 

As the first step of SP-based methods, question understanding module converts unstructured text into encoded question, which benefits the downstream parsing. 
Compared with simple questions, complex questions are featured with compositional semantics and more complex query types, which increase the difficulty in linguistic analysis. 

\subsubsection{Understanding complex semantics of questions}

The complex semantics of questions indicates a complex dependency pattern of sentences, which expresses the relation between constituents.
Knowing the core part of the sentence structure could be beneficial for question understanding.
Incorporating structure property of questions is an intuitive strategy to achieve this goal.


\paratitle{Incorporating structure property of questions to seq2seq generation.}
Many existing methods rely on syntactic parsing, such as dependencies~\cite{Abujabal-WWW-17,Abujabal-WWW-2018,Luo-EMNLP-2018} and Abstract Meaning Representation~(AMR)~\cite{Kapanipathi-AAAI-2021}, to provide better alignment between question constituents and logic form elements (\eg entity, relation, entity types, and attributes). This line of research is illustrated at the left side of Figure~\ref{fig-syntactic}. 
In order to represent long-range dependencies between the answer and the topic entity in question, Luo \etal~\cite{Luo-EMNLP-2018} extracted the dependency path between them. By encoding the directional dependency path, they concatenated both syntactic features and local semantic features together to form global question representation.
Similarly, Abujabal \etal~\cite{Abujabal-WWW-17} leveraged dependency parse to cope with compositional utterances and only focused on important tokens contained by parsed dependency path when creating query templates.
Instead of directly creating logic forms upon the dependency paths, Abujabal \etal~\cite{Abujabal-WWW-2018} leveraged dependency parse to analyze the composition of the utterances and aligned it with the logic form. 
Kapanipathi \etal~\cite{Kapanipathi-AAAI-2021} introduced AMR to help understand questions, the benefits are two-fold: (1) AMR is effective in disambiguating the natural language utterances. (2) AMR parsing module is highly abstract and helps to understand the questions in a KB-independent way. 
However, the accuracy of producing syntactic parsing is still not satisfying on complex questions, especially for those with long-distance dependency. 

In order to alleviate the inaccurate syntactic parsing of complex questions, Sun \etal~\cite{Skeleton-AAAI-2020} leveraged the skeleton-based parsing to obtain the trunk of a complex question, which is a simple question with several branches (\ie head word of original text-spans) to be expanded. 
For example, the trunk for question ``\textit{What movie that Miley Cyrus acted in had a director named Tom Vaughan ?}'' is ``\textit{What movie had a director ?}'', and attributive clauses in question will be regarded as the branches of the trunk. 
Under such a skeleton structure, only simple questions are to be parsed further, which is more likely to obtain accurate parsing results.

\subsubsection{Understanding complex syntax of queries}

It is important to understand questions by analyzing their complex semantics.
It is also crucial to analyze the syntax of queries and ensure that the generated logic forms could meet the complex syntax of queries.
While above methods generate logic forms with Seq2seq framework, another line of work (shown at the right side of Figure~\ref{fig-syntactic}) focuses on leveraging structural properties (\eg tree structure or graph structure of logic forms) for ranking candidate parsing. 

\paratitle{Incorporating structure property of logic forms to feature-based ranking.}
Maheshwari \etal~\cite{Maheshwari-ISWC-2019} proposed a novel ranking model which exploits the structure of query graphs and uses attention weights to explicitly compare predicates with natural language questions. 
Specifically, they proposed a fine-grained slot matching mechanism to conduct hopwise semantic matching between the question and each predicate in the core reasoning chain. 
Instead of capturing semantic correlations between a question and a simple relation chain, Zhu \etal~\cite{Zhu-Neuro-2020} focused on structure properties of query and conducted KBQA with query-question matching. They employed a structure-aware encoder to model entity or relation context in a query, promoting the matching between queries and questions. 
Similarly, Zafar \etal~\cite{Zafar-ESWC-2018} incorporated two Tree-LSTMs~\cite{Tai-ACL-2015} to model dependency parse trees of questions and tree structure of candidate queries respectively, and leveraged structural similarity between them for comprehensive ranking.

Traditional methods adopted state-transition strategy to generate candidate query graphs. As this strategy ignores the structure of queries, a considerable number of invalid queries will be generated as candidates. To filter these queries out, Chen \etal~\cite{Chen-IJCAI-2020} proposed to predict the query structure of the question and leverage the structure to restrict the generation of the candidate queries.
Specifically, they designed a series of operations to generate placeholders for types, numerical operators, predicates, and entities. After that, they can ground such uninstantiated logic forms with KBs and generate executable logic forms. 


\subsection{Parsing Complex Queries}

\begin{figure}[t!]
    \centering
    \includegraphics[scale=0.47]{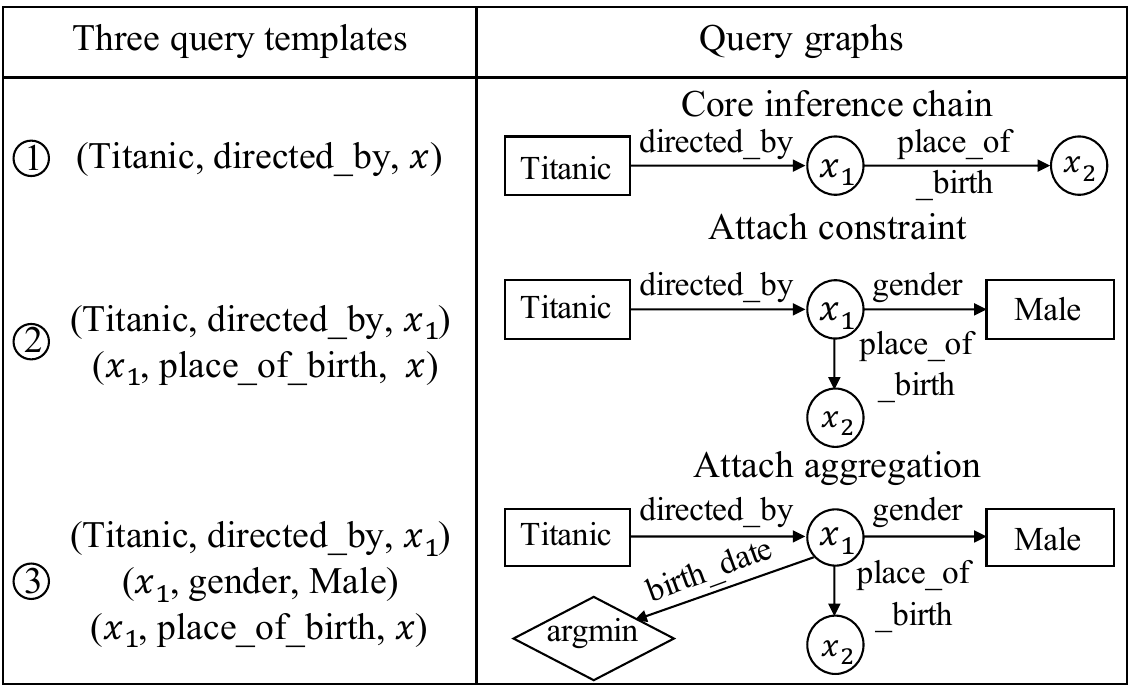}
    \caption{The illustration of possible parsing targets of for a complex question ``\textit{What's the birth place of the youngest male director of Titanic?}''.
    $x$ denotes the satisfied entity to be queried and $x_1$ denotes the intermediate entity included in the query.}
    \label{fig:parsing_target}
\end{figure}


To generate an executable logic form, traditional methods first utilized the existing parsers to convert a question into CCG derivation which is then mapped to a SPARQL via aligning predicates and arguments to relations and entities in the KBs~\cite{Cai-ACL-2013}. Such methods are sub-optimal for complex questions due to the ontology mismatching problem~\cite{Kwiatkowski-EMNLP-2013}.
Thus it is necessary to leverage the structure of KBs for accurate parsing, where parsing is performed along with the grounding of the KB. 

\paratitle{Designing logic forms via pre-defined query templates.}
To satisfy the compositionality of the complex questions, researchers have developed diverse expressive logic forms as parsing targets.
Recall the topic entities are recognized in the preliminary step, Bast \etal~\cite{Bast-CIKM-15} started from the topic entities and designed three query templates as the parsing targets. 
We list these three query templates in Figure~\ref{fig:parsing_target}.
The first two templates return entities which are 1-hop and 2-hop away from the topic entities ``\textit{Titanic}''.
The third template returns entities that are two hops away from the topic entities and constrained by another entity.
A follow-up study concentrated on designing templates to answer temporal questions~\cite{Jia-CIKM-2018}. 
Although such template-based methods can successfully parse several types of complex questions, it suffers from the limited coverage issue.

\paratitle{Designing expressive logic forms with flexible combining rules.}
To design more expressive logic forms, Yih \etal~\cite{Yih-ACL-2015} proposed query graph as the expressive parsing target.
A query graph is a logic form in graph structure which closely matches with the KB schemas and is an alternative to an executable SPARQL. It consists of entities, variables, and functions, which correspond to grounded entities mentioned in questions, variables to query and aggregation operations, respectively. 
As illustrated in Figure~\ref{fig:parsing_target}, a set of core inference chains~\cite{Yih-ACL-2015} starting from the topic entity are first identified.
Constraint entities and aggregation operators are further attached to the path chains to make them adapt to more complex questions.
Unlike the pre-defined templates, query graphs are not limited to the hop and constraint numbers. 
They have shown strong capabilities to express complex questions while they are still incapable to deal with long-tail complex question types.
Based on more observations towards the  long-tail data samples, follow-up work tried to improve the formulation of query graphs by involving syntactic annotation to enhance the structural complexity of the query graph~\cite{Abujabal-WWW-17}, applying more aggregation operators such as merging, coreference resolution~\cite{Hu-EMNLP-18} to fit complex questions. 
Compared with query templates, logic forms with flexible combining rules could fit into a large variety of complex queries.
A more expressive logic form indicates a more robust KBQA system which can handle questions with greater diversity.

\subsection{Grounding with Large Search Space}

\begin{figure}[t!]
    \centering
    \includegraphics[scale=0.453]{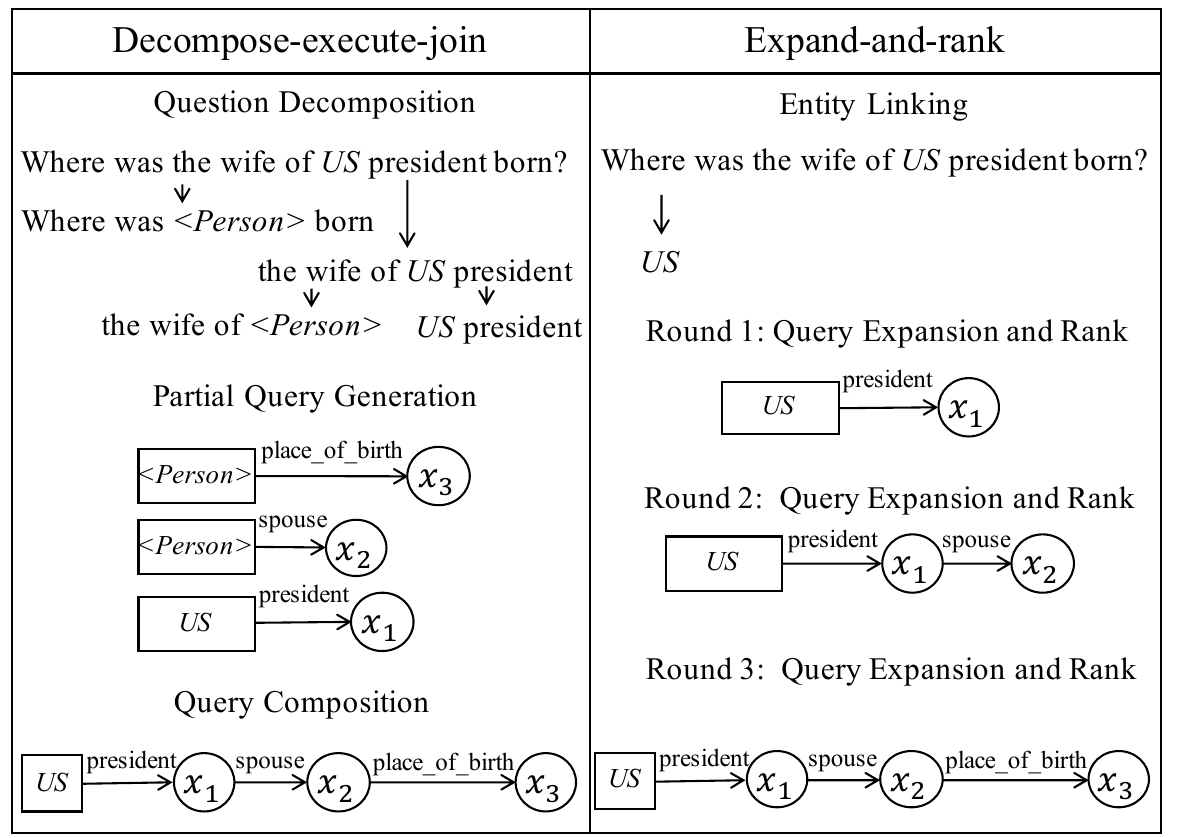}
    \caption{Illustration of two grounding strategies for a complex question ``\textit{Where was the wife of US president born?}''.
    }
    \label{fig:grounding_space}
\end{figure}
To obtain executable logic forms, KB grounding module instantiates possible logic forms with a KB.
As one entity in the KB can be linked to hundreds or even thousands of relations, it is unaffordable to explore and ground all the possible logic forms for a complex question considering both computational resource and time complexity.

\paratitle{Decomposing a complex question to sub-questions.}
Instead of enumerating logic forms with a single pass, researchers try to propose methods to generate the complex queries with multiple steps.
Zheng \etal~\cite{Zheng-VLDB-2018} proposed to first decompose a complex question into multiple simple questions, where each simple question was parsed into a simple logic form. 
The final answers are obtained with either the conjunction or composition of the partial logic forms.
This \textbf{\textit{decompose-execute-join}} strategy can effectively narrow down the search space.
A similar approach was studied by Bhutani \etal~\cite{Bhutani-CIKM-2019}. 
As decomposing questions costs manual efforts, they reduced human annotation and identify the composition plan through an augmented pointer network~\cite{CWQ-NAACL-2018}. 
The final answers are obtained via conjunction or composition of the answers of decomposed questions.

\paratitle{Expanding a logic form by iteration.}
Unlike decomposing a complex question to sub-questions, a number of studies adopted the \textbf{\textit{expand-and-rank}} strategies to reduce the search space by expanding the logic forms in an iterative manner.
Specifically, they collected all the query graphs that are 1-hop neighborhood of the topic entities as the candidate logic forms at the first iteration.
These candidates are ranked based on their semantic similarities with the question.
Top-ranked candidates are kept to do further expansion while low-ranked candidates are filtered out.
At the following iterations, each top-ranked query graph in the beam is extended, which results in a new set of candidate query graphs that are more complicated.
This procedure will repeat until the best query graph is obtained.
Chen \etal~\cite{Chen-NAACL-2019} first utilized the hopwise greedy search to expand the most-likely query graphs.
Lan \etal~\cite{Lan-ICDM-2019} proposed an incremental sequence matching module to iteratively parse the questions without revisiting the generated query graphs at each searching step.
Above expansion is conducted in a linear manner, which is only effective in generating multi-hop relations.
Lan \etal~\cite{Lan-ACL-2020} defined three expansion actions for each iteration, which are extending, connecting, and aggregating to correspond to multi-hop reasoning, constrained relations, and numerical operations, respectively.
Examples in Figure~\ref{fig:grounding_space} show the different principles of these two strategies.

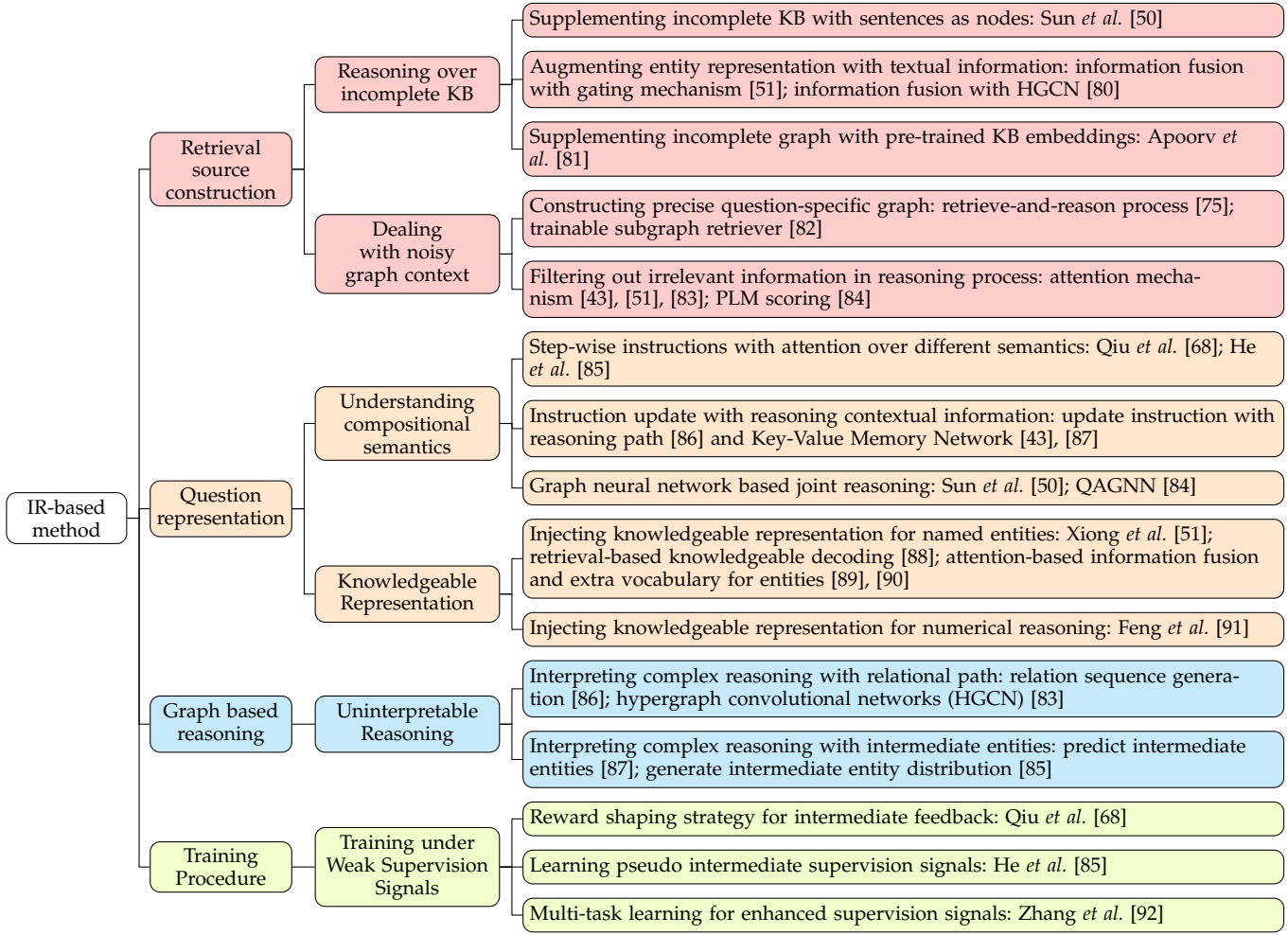
\begin{figure*}
	\footnotesize
	\begin{forest}
		for tree={
			forked edges,
			grow'=0,
			draw,
			rounded corners,
			node options={align=center},
			calign=edge midpoint,
		},
	    [IR-based method, text width=1.5cm, for tree={fill=white!20}
			[Retrieval source construction, text width=1.8cm, for tree={fill=red!20}
				[Reasoning over incomplete KB, text width=2.4cm, for tree={fill=red!20}
				    [
				    Supplementing incomplete KB with sentences as nodes: Sun \etal~\cite{Sun-EMNLP-2018},
				    text width=10.5cm, node options={align=left}
				    ]
				    [
				    Augmenting entity representation with textual information: information fusion with gating mechanism~\cite{Xiong-ACL-2019}; information fusion with  HGCN~\cite{Han-EMNLP-2020},
				    text width=10.5cm, node options={align=left}
				    ]
				    [
				    Supplementing incomplete graph with pre-trained KB embeddings: Apoorv \etal~\cite{Apoorv-ACL-2020},
				    text width=10.5cm, node options={align=left}
				    ]
				]
				[Dealing with noisy graph context, text width=2.4cm, for tree={fill=red!20}
					[
					Constructing precise question-specific graph: retrieve-and-reason process~\cite{Sun-EMNLP-2019}; trainable subgraph retriever~\cite{Zhang-arxiv-2022},
					text width=10.5cm, node options={align=left}
					]
					[
					Filtering out irrelevant information in reasoning process: attention mechanism~\cite{KVMem-EMNLP-2016,Xiong-ACL-2019,Han-IJCAI-2020}; PLM scoring~\cite{Yasunaga-NAACL-2021},
					text width=10.5cm, node options={align=left}
					]
				]
			]
			[Question representation, text width=1.8cm, for tree={fill=orange!20}
				[Understanding compositional semantics, text width=2.4cm, for tree={fill=orange!20}
					[
					Step-wise instructions with  attention over different  semantics: Qiu \etal~\cite{Qiu-WSDM-2020}; He \etal~\cite{He-WSDM-2021},
					text width=10.5cm, node options={align=left}
					]
					[
					Instruction update with reasoning contextual information: update instruction with reasoning path~\cite{Zhou-COLING-2018} and Key-Value Memory Network~\cite{KVMem-EMNLP-2016,Xu-NAACL-2019},
					text width=10.5cm, node options={align=left}
					]
					[
					Graph  neural  network  based  joint  reasoning: Sun \etal ~\cite{Sun-EMNLP-2018}; QAGNN~\cite{Yasunaga-NAACL-2021},
					text width=10.5cm, node options={align=left}
					]
				]
				[Knowledgeable Representation, text width=2.4cm, for tree={fill=orange!20}
					[
					Injecting  knowledgeable  representation for named  entities: Xiong \etal~\cite{Xiong-ACL-2019}; retrieval-based knowledgeable decoding~\cite{He-ACL-2017}; attention-based information fusion and extra vocabulary for entities~\cite{Yin-IJCAI-2016,Fu-NAACL-2018}
					,text width=10.5cm, node options={align=left}
					]
					[
					Injecting  knowledgeable  representation for numerical reasoning: Feng \etal~\cite{Feng-EMNLP-2021},
					text width=10.5cm, node options={align=left}
					]
				]
			]
			[Graph based reasoning, text width=1.8cm, for tree={fill=cyan!20}
				[Uninterpretable Reasoning, text width=2.4cm, for tree={fill=cyan!20}
					[
					Interpreting complex reasoning with relational path: relation sequence generation~\cite{Zhou-COLING-2018}; hypergraph convolutional networks (HGCN)~\cite{Han-IJCAI-2020},
					text width=10.5cm, node options={align=left}
					]
					[
					Interpreting complex reasoning with intermediate entities: predict intermediate entities~\cite{Xu-NAACL-2019}; generate intermediate entity distribution~\cite{He-WSDM-2021},
					text width=10.5cm, node options={align=left}
					]
				]
			]
			[Training Procedure, text width=1.8cm, for tree={fill=lime!20}
				[Training under Weak Supervision Signals, text width=2.4cm, for tree={fill=lime!20}
					[
					Reward shaping strategy for intermediate feedback: Qiu \etal~\cite{Qiu-WSDM-2020},
					text width=10.5cm, node options={align=left} 
					]
					[
					Learning pseudo intermediate supervision signals: He \etal~\cite{He-WSDM-2021},
					text width=10.5cm, node options={align=left}
					]
					[
					Multi-task learning for enhanced supervision signals: Zhang~\etal~\cite{Zhang-AAAI-2018},
					text width=10.5cm, node options={align=left}
					]
				]
			]
		]
	\end{forest}
	\caption{The main content of IR-based methods. The hierarchical structure is arranged with: IR-based method $\rightarrow$ module $\rightarrow$ challenge  $\rightarrow$ solution.}
    \label{fig:taxonomy-ir}
\end{figure*}

\subsection{Training under Weak Supervision Signals}
To cope with the issue of unlabeled reasoning paths, reinforcement learning (RL) based optimization has been used to maximize the expected reward~\cite{Liang-ACL-17,Qiu-CIKM-2020}.
However, the insufficient training data makes it a challenge to train under weak supervision.

\subsubsection{Training with sparse reward}
Training via RL indicates that SP-based methods can only receive the feedback after the execution of the complete parsed logic form.
This leads to a long exploration stage with severe sparse positive reward.
To tackle this issue, methods are proposed to augment the final reward or intermediate reward. 

\paratitle{Augmenting final reward with enriched features.}
Some research work adopted reward shaping strategy for parsing evaluation.
Specifically, researchers augment reward of a logic form by involving more information of answers as the enriched features of the final prediction.
Saha \etal~\cite{Saha-TACL-2019} rewarded the model by the additional feedback when the predicted answers have the same type as ground truth. 
In this way, even the predicted answers are not  exactly the ground truth, they could also encourage the model to search for the right answer type. 
This helps to avoid the sparse positive rewards during the exploration stage.

\paratitle{Augmenting intermediate reward with enriched critics.}
Besides rewards derived from the final prediction, intermediate rewards during the semantic parsing process may also help address this challenge. 
Recently, Qiu \etal~\cite{Qiu-CIKM-2020} formulated query graph generation as a hierarchical decision problem, and proposed an option-based hierarchical framework to provide intermediate rewards for low-level agents. Through options over the decision process, the high-level agent sets goal for the low-level agent at intermediate steps. 
To evaluate whether intermediate states of the low-level agent meet the goal of the high-level agent, they measured the semantic similarity between the given question and the generated triple. To provide the policy with effective intermediate feedback, Qiu \etal~\cite{Qiu-CIKM-2020} augmented the critic of query graphs with hand-crafted rules.


\subsubsection{Dealing with spurious reasoning}
At the early stage of training, it is difficult to find a logic form with the positive reward. 
Moreover, random exploration at the early stage easily lead to spurious reasoning, where logic forms result in correct answers but are semantically incorrect.
Therefore, the early supervision of high-quality logic forms could be conducted to speed up the training and prevent models from the misguiding of spurious reasoning.

\paratitle{Stabilizing training processing with high-reward logic forms.}
To accelerate and stabilize the training process, Liang \etal~\cite{Liang-ACL-17} proposed to maintain pseudo-gold programs found by an iterative maximum-likelihood training process to bootstrap training. The training process contains two steps: (1) leveraging beam search mechanism to find pseudo-gold programs, and (2) optimizing the model under the supervision of the best program found in history. 
Hua \etal~\cite{Hua-JWS-2020} followed a similar idea to evaluate the generated logic form by comparing it with the high-reward logic forms stored in the memory buffer. To make a trade-off between exploitation and exploration,  they proposed the proximity reward and the novelty reward to encourage remembering the past high-reward logic forms and generating new logic forms to alleviate spurious reasoning respectively. 
Combining such bonus with terminal reward, models can obtain dense feedback along the learning phrase.

%% file: sec-ir.tex
\section{Information Retrieval-based Methods}
\label{sec-IR}
\ignore{
As mentioned in Sec.~\ref{sec-methods}, information retrieval-based methods generally perform multi-hop reasoning over the question-specific graph to find the answer.
Information retrieval-based methods first construct the question-specific graph as either the entire KBs or the partial subgraph, which face the \textbf{incomplete knowledge} owing to the incompleteness of KBs.
And then, they need to perform multi-hop reasoning over the question-specific graph.
In order to analyse and optimize the answer inference for complex KBQA, the intermediate reasoning status is supposed to be explicitly visible.
However, existing methods directly viewed the whole reasoning process as vector similarity matching, which caused the \textbf{lack of interpretability}.
Similar as the semantic parsing-based methods, this category also encounters the \textbf{weak supervision signal} when training, owing to the lack of ground truth annotation of multi-hop reasoning path.
The following part illustrate how prior work deal with these challenges.
}
In this section, we summarize the main challenges brought by complex questions for different modules of IR-based methods. The taxonomy of challenges and solutions can be visualized with Figure~\ref{fig:taxonomy-ir}.

\subsection{Overview}
The overall procedure typically consists of the modules of retrieval source construction, question representation, graph based reasoning, and answer generation.  
These modules will encounter different challenges for complex KBQA. 
Firstly, the retrieval source module extracts a question-specific graph from KBs, 
which includes both relevant facts and a wide range of noisy facts. Due to unneglectable incompleteness of source KBs~\cite{Min-NAACL-2013}, the correct reasoning paths may be absent from the extracted graph. The two issues are more likely to occur in the case of complex questions. 
Secondly, question representation module understands the question and generates instructions to guide the reasoning process. 
This step becomes challenging when the question is complicated. 
After that, reasoning on graph is conducted through semantic matching. 
When dealing with complex questions, such methods rank answers through semantic similarity without traceable reasoning in the graph, which hinders reasoning analysis and failure diagnosis.

The following parts illustrate how prior work deals with these challenges and the utilized advanced techniques.

\subsection{Reasoning under Imperfect KB}
\ignore{
}
In general, IR-based methods find answers by conducting reasoning on a graph structure.
This graph structure is a question-specific graph extracted from a KB in most cases. However, such question-specific graphs are never perfect, due to incompleteness of KBs and the noisy graph context brought by heuristic graph generation strategy.

\subsubsection{Reasoning over incomplete KB}
\begin{figure*}[htbp]
    \centering
    \includegraphics[scale=0.58]{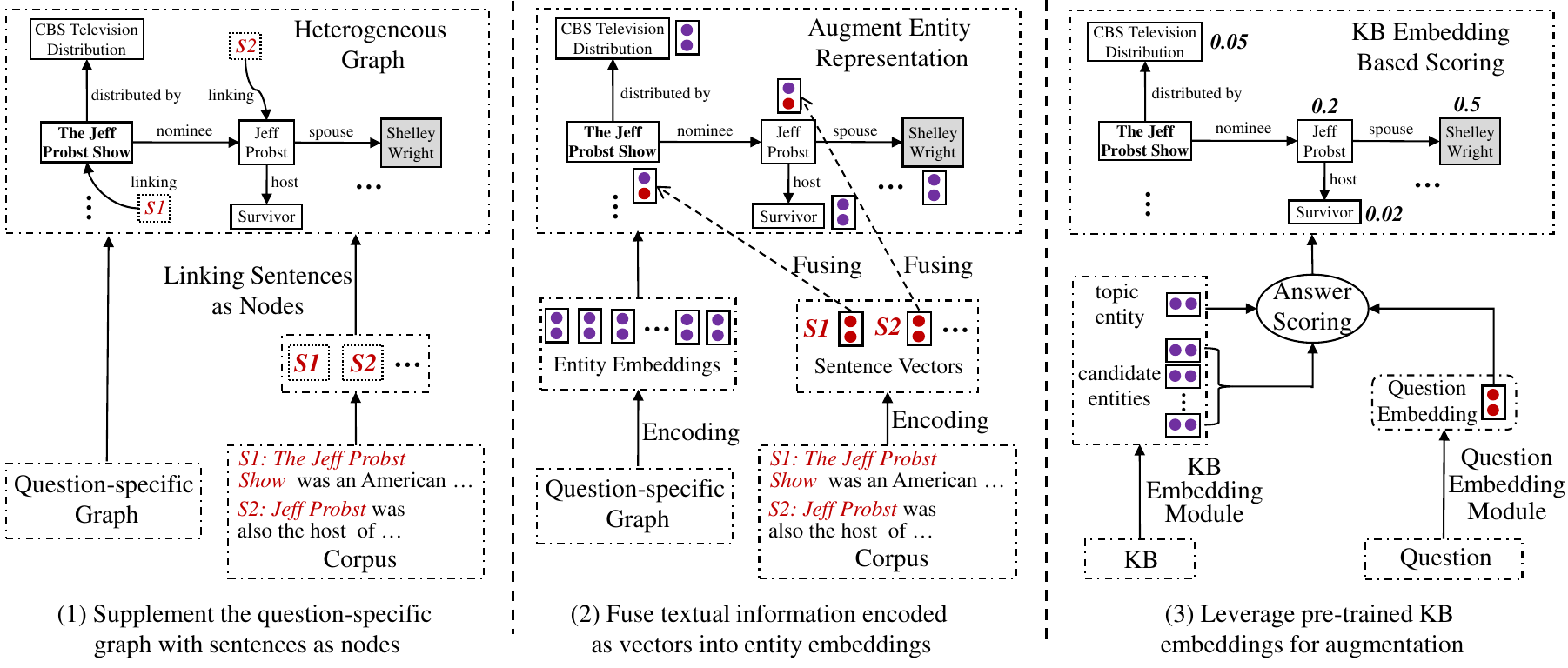}
    \caption{Illustration of three categories of methods to supplement the incomplete KB. 
    All subfigures are drawn in a bottom-up style, where the input is placed at the bottom and supplemented graph is placed on the top. 
    The topic entity and the answer entity are shown in the bold font and shaded box respectively. 
    }
    \label{fig:supplement}
\end{figure*}

It is vital for the question-specific graph to obtain high recall of correct reasoning paths. 
Since simple questions only require 1-hop reasoning on the neighborhood of the topic entity in the KB, IR-based methods are less likely to suffer from the inherent incompleteness of KBs~\cite{Min-NAACL-2013} when solving simple questions. 
By contrast, the correct reasoning paths for complex questions are of high probability to be absent from the question-specific graph and it turns to be a severe issue. 
To tackle with this challenge, researchers utilize auxiliary information to supplement the knowledge source.
We divide the different supplementary methods into three categories and show the core differences in Figure~\ref{fig:supplement}.

\paratitle{Supplementing incomplete KB with sentences as nodes.} Intuitively, a large amount of question-relevant text corpus retrieved from Wikipedia can provide a wide range of unstructured knowledge as supplementary evidence. Based on this observation, Sun \etal~\cite{Sun-EMNLP-2018} proposed to complement the graph with extra question-relevant  sentences as nodes and reason on the augmented heterogeneous graph (\ie the left side of Figure~\ref{fig:supplement}). According to the entities mentioned in sentences, they linked them to corresponding entities on the graph and viewed them as nodes.

\paratitle{Augmenting entity representation with textual information.} Instead of directly complementing sentences to the question-specific graph as nodes, Xiong \etal~\cite{Xiong-ACL-2019} and Han \etal~\cite{Han-EMNLP-2020} proposed to fuse extra textual information into the entity representation as the second way (shown in the middle of Figure~\ref{fig:supplement}). 
Xiong \etal~\cite{Xiong-ACL-2019} designed a novel conditional gating mechanism to obtain knowledge-aware information of sentences under the guidance of text-linked entity representations extracted with a subgraph reader.
Such knowledge-aware information of sentences is further aggregated to enhance the entity representations to complement incomplete KB. 
Similarly, Han \etal~\cite{Han-EMNLP-2020} fused textual information of sentences into entity representations. 
In their settings, every sentence is regarded as a hyperedge connecting all of its involved entities, and a document can be viewed as a hypergraph. 
Based on hypergraph convolutional networks~(HGCN)~\cite{Feng-AAAI-2019}, they encoded the sentences in the document and fused sentence representations into sentence-linked entity representations.

\paratitle{Supplementing incomplete graph with pre-trained KB embeddings.} In knowledge base completion (KBC) task, knowledge base embeddings have been adopted to alleviate the sparsity of KB by performing missing link prediction. 
Inspired by that, Apoorv \etal~\cite{Apoorv-ACL-2020} utilized pre-trained knowledge base embeddings to address the incomplete KB issue as shown at the right side of Figure~\ref{fig:supplement}.
Specifically, they pre-trained KB embeddings (\ie entity and relation embeddings) with ComplEX~\cite{Trouillon-complexe-2016} approach 
and predicted the answer via a triple scoring function taking the triples in the format of (topic entity, question, answer entity) as inputs. 
To make questions fit into original ComplEX scoring function, they map Roberta~\cite{Liu-roberta-2019} embeddings of question into the complex space of same dimension. 
By leveraging the pre-trained knowledge from global KBs, they implicitly complemented the incomplete question-specific graph.

\subsubsection{Dealing with noisy graph context}
Since question-specific graphs are always constructed with heuristics~\cite{Sun-EMNLP-2018}, it may introduce redundant and even question-irrelevant noisy graph context (both entities and sentence nodes). Compared with simple questions which require only 1-hop reasoning, the question-specific graphs constructed for complex questions are more likely to involve noisy graph context.
Reasoning over such noisy graphs poses a great challenge for complex questions, meanwhile it also reduces the efficiency of model training.

\paratitle{Constructing precise question-specific graph.} 
An intuitive idea is to construct a relatively small and precise graph for downstream reasoning. 
To achieve this goal, Sun \etal~\cite{Sun-EMNLP-2019} proposed to build the heterogeneous graph with an iterative retrieve-and-reason process under the supervision of shortest paths between the topic entities and answer entities. In a recent work, Zhang \etal~\cite{Zhang-arxiv-2022} proposed a trainable subgraph retriever (SR) which retrieves relevant relational paths for subsequent reasoning. And their experimental results proved such precise graphs can bring substantial performance gains for IR-based methods. 

\paratitle{Filtering out irrelevant information in reasoning process.} Besides constructing small and precise graphs for subsequent reasoning, some research work proposed to filter irrelevant information out along the reasoning process. Attention mechanisms, which are effective in eliminating irrelevant features, have been adopted by existing IR-based methods~\cite{KVMem-EMNLP-2016,Xiong-ACL-2019,Han-IJCAI-2020} to reserve relevant information during the reasoning process. 
Similarly, Yasunaga \etal~\cite{Yasunaga-NAACL-2021} adopted pre-trained language model scoring of each node conditioned on question answering context as relevance scores to guide subsequent reasoning process.

\subsection{Understanding Complex Semantics}
Understanding the complex questions is the prerequisite for subsequent reasoning. However, complex questions contain compositional semantics and require specific knowledge (\eg named entities, ordinal reasoning) to answer. Due to such intrinsic properties of complex questions, methods designed for simple question understanding may be not fit for complex questions.

\subsubsection{Understanding compositional semantics}
IR-based methods usually generate initial question representation $\bm{q}$  by directly encoding questions as low-dimensional vectors through neural networks (\eg LSTM and GRU).
Static reasoning instruction (\eg final hidden states of $\bm{q}$) obtained through above approach can not effectively represent the compositional semantics of complex questions, which poses challenges to guide the reasoning over the question-specific graph.
\ignore{
}
In order to comprehensively understand questions, some studies dynamically update the reasoning instruction during the reasoning process.
\paratitle{Step-wise instructions with attention over different semantics.} To make the reasoning models aware of the reasoning step, Qiu \etal~\cite{Qiu-WSDM-2020} proposed to learn a step-aware representation through transforming initial question representation $\bm{q}$ with a single-layer perceptron. 
After obtaining step-aware question representation, attention mechanism is further incorporated to select useful information to generate instruction vectors. 
Similarly, He \etal~\cite{He-WSDM-2021} proposed to focus on different parts of the question with dynamic attention mechanism.
Based on both step-aware question representation and previous reasoning instruction $\bm{i}^{(k-1)}$, they generated attention distribution over tokens of the question and updated the instruction vector. 


\paratitle{Instruction update with reasoning contextual information.} 
Besides explicitly recording the analyzed part of the question via attention, some other work proposed to update the instruction with information retrieved along the reasoning process. 
A typical example is generating explicit reasoning paths and updating  instruction with generated paths. 
Zhou \etal~\cite{Zhou-COLING-2018} designed a model that takes the current reasoning instruction $\bm{i}^{(k)}$ as the input, and then predicts the intermediate relation $r^{(k)}$ from all relations in KB.
After obtaining the predicted relation, 
the model updated the instruction vector as: $\bm{i}^{(k+1)} = \bm{i}^{(k)} - \bm{r}^{(k)}$, where the subtraction is meant to omit the analyzed information from the question.
Thus, the updated reasoning instruction can hold unanalyzed parts of the question in the subsequent reasoning process.

Instead of generating explicit reasoning paths, Xu~\etal~\cite{Xu-NAACL-2019} and Miller~\etal~\cite{KVMem-EMNLP-2016} employed key-value memory network to achieve similar dynamic instruction update.
Specifically, they first included all KB facts that contains with one of topic entities as subject into the memory.
Then, they indexed the keys and values in the key-value memory, where keys are (subject, relation) pairs and values are corresponding object entities. 
A key addressing process is conducted to find the most suitable key and corresponding value for the instruction. 
With the addressed key and value, they concatenated their representations with the previous step reasoning instruction $\bm{i}^{(k)}$ and performed a linear transformation to obtain the updated reasoning instruction $\bm{i}^{(k+1)}$ to guide the next hop reasoning.
In this way, the reasoning instruction will be updated over the memory.

\paratitle{Graph neural network based joint reasoning.}
Besides instruction update, another line of research addresses such compositional semantics with graph neural network~(GNN) based reasoning. Sun \etal~\cite{Sun-EMNLP-2018}  proposed a GNN-based model GraftNet to reason complex questions over heterogeneous information sources. Through iterative GNN reasoning steps, the entity representations and reasoning instruction get updated in turn. The reasoning instruction conveys the knowledge of the topic entity which is dynamically updated over the reasoning process.
\ignore{
At each step, they updated the representation of entity nodes and text nodes by performing propagation and aggregation in the heterogeneous graph.
After that, they computed the reasoning instruction $\bm{i}^{(k+1)}$ through aggregating $\bm{i}^{(k)}$ with the representation of entities which are topic entities of the question. 
The reasoning instruction conveys the knowledge of the topic entity which is dynamically updated over the reasoning process. }
Despite iterative update of reasoning instruction and graph neural network, Yasunaga \etal proposed~\cite{Yasunaga-NAACL-2021} QAGNN model which reasoned complex questions with single graph neural network based joint reasoning. They constructed the question-specific graph with an extra question-answering context node which connects with all other nodes in the graph. All nodes are uniformly encoded with pre-trained language models (PLMs) as initial representation, and get updated along with graph neural network reasoning.

\subsubsection{Knowledgeable Representation}
Apart from compositional semantics, complex questions may also contain knowledge-intensive tokens or phrases (\eg named entities, ordinal constraint), which hinders natural language understanding for text-based semantic understanding. Besides question text, external knowledge is taken as the input to help understand these complex questions.

\paratitle{Injecting knowledgeable representation for named entities.} 
In the natural language questions, the topic entities are always named entities which are not informative enough for understanding. To cope with such named entities, some existing work proposed to inject more informative representations obtained from knowledge bases. As a typical example, Xiong \etal~\cite{Xiong-ACL-2019} proposed to reformulate query representation in latent space with knowledge representation learned from the graph context of topic entities. Through ablation study, they verified the effectiveness of injecting such knowledgeable representation into question representation. Similar ideas were also adopted in knowledge-enhanced language model pre-training~\cite{ERNIE-ACL-2019,Peters-EMNLP-2019}.

While natural answers can be generated from popular seq2seq text generation framework, it is still hard to directly generate the named entities from token vocabulary. To address this gap, He \etal~\cite{He-ACL-2017} first proposed a copying and retrieving mechanism to generate the natural answers from extra vocabulary for question tokens and entities in the question-specific graph. Similarly, Yin \etal ~\cite{Yin-IJCAI-2016} and Fu \etal~\cite{Fu-NAACL-2018} fed relational facts into structured memory slots, which served as extra vocabulary to generate named entities, and generate knowledgeable representation with attention-based information fusion.

\paratitle{Injecting knowledgeable representation for numerical reasoning.}
While multiple solutions are proposed to conduct multi-hop reasoning, little attention is paid to solving complex questions with numerical operations. 
To empower IR-based methods with numerical reasoning capability, Feng~\etal~\cite{Feng-EMNLP-2021} proposed to encode numerical properties (\ie the magnitude and ordinal properties of numbers) into entity representations. First, they manually defined a list of ordinal determiners (\eg first, largest) to detect ordinal constrained questions. For these detected questions, they enrich their question-specific graphs with extra numerical attribute triplets. Encoding these numerical attribute triplets with pre-trained number encoding modules, extra number embeddings can be used as model-agnostic plugins to conduct numerical reasoning for IR-based methods.

\subsection{Uninterpretable Reasoning}
Since the complex questions usually query multiple facts in sequence, the system is supposed to accurately predict answers over the graph based on a traceable reasoning process. 
While neural networks are powerful, blackbox style of the reasoning module makes the reasoning process less interpretable and hard to incorporate user interaction for further improvement. 
To derive a more interpretable reasoning process, the reasoning is  performed with a multi-step  intermediate prediction. 
Along the reasoning process, the KBQA model generates a series of reasoning status $\{s^{(k)}, k=1, ..., n\}$. 
While the final status is leveraged to generate the answer prediction, the intermediate status may help generate intermediate predictions (\ie matched relations or entities) for better interpretability. 
More importantly, intermediate predictions make detecting spurious or error reasoning easier with user interaction.

\paratitle{Interpreting complex reasoning with relational path.} Existing studies adopted different designs of reasoning status and reasoning modules to interpret the reasoning process.
Specifically, Zhou \etal~\cite{Zhou-COLING-2018} formulated the multi-hop reasoning process as relation sequence generation and represented reasoning status using a vector. 
For each step, instruction vector and status vector are matched with relation candidates to generate probability distribution over all relations in KB. And weighted relation representation is then leveraged to update the status. By repeating this process, the model can achieve an interpretable reasoning process. 
Inspired by above work, Han \etal~\cite{Han-IJCAI-2020} proposed an interpretable model based on hypergraph convolutional networks (HGCN) to predict relation paths for explanation.
They constructed a dense hypergraph by pinpointing a group of entities connected via same relation, which simulated human’s hopwise relational reasoning. 
To train these two models, gold relation paths are leveraged. However, gold relation path annotations are unavailable in most cases, which makes their methods inapplicable to general datasets.

\paratitle{Interpreting complex reasoning with intermediate entities.} Apart from relation paths, some research work predicted  question-relevant entities at intermediate steps to explain multi-hop reasoning process. 
Xu \etal~\cite{Xu-NAACL-2019} elaborately adopted key-value memory network to achieve a traceable reasoning process.
In their work, status $s^{(k)}$ is defined as the weighed sum of value representation, the weight of which is derived from key-instruction matching. 
To predict intermediate entities, their model  followed traditional IR-based methods to score candidates given query 
$s^{(k)} + s^{(k-1)}$.
As spurious long paths may connect topic entities with answer entities in KB, during training, they proposed to supervise intermediate entity prediction with the final answers. 
Such objective encourages the model to generate shortest reasoning path. 
Besides explicitly generating intermediate entities, He \etal~\cite{He-WSDM-2021} proposed to generate intermediate entity distribution to indicate the reasoning process. Their experimental results also showed that such intermediate supervision signals can effectively reduce spurious reasoning.

\subsection{Training under Weak Supervision Signals} 
Similar to the SP-based methods, it is difficult for IR-based methods to reason the correct answers without any annotations at intermediate steps, since the model cannot receive any feedback until the end of reasoning. It is found that this case may lead to spurious reasoning~\cite{He-WSDM-2021}. 
Due to the lack of intermediate state supervision signals, the reward obtained from spurious reasoning may mislead the model. 

\paratitle{Reward shaping strategy for intermediate feedback.} To train model under weak supervision signals, Qiu \etal~\cite{Qiu-WSDM-2020} formulated multi-hop reasoning process over KBs as a process of expanding the reasoning path on graph. Based on the encoded decision history, the policy network leveraged attention mechanism to focus on the unique impact of different parts of a given question over triple selection.
To alleviate the delayed and sparse reward problem caused by weak supervision signals, they adopted reward shaping strategy to evaluate reasoning paths and provide intermediate rewards.
Specifically, they utilized semantic similarity between the question and the relation path to evaluate reasoning status at intermediate steps.

\paratitle{Learning pseudo intermediate supervision signals.} Besides evaluating the reasoning status at intermediate steps, a more intuitive idea is to infer pseudo intermediate status and augment model training with such inferred signals. 
Inspired by bidirectional search algorithm on graph, He \etal~\cite{He-WSDM-2021} proposed to learn and augment intermediate supervision signals with bidirectional reasoning process. 
Taking entity distribution as suitable supervision signals at intermediate steps, they proposed to learn and leverage such signals under teacher-student framework. 

\paratitle{Multi-task learning for enhanced supervision signals.} While most of existing work focused on enhancing the supervision signals at intermediate steps, few work paid attentions to the entity linking step.
Most of existing work utilized off-the-shelf tools to locate the topic entity in question, causing error propagation.
In order to accurately locate the topic entity without annotations, Zhang \etal~\cite{Zhang-AAAI-2018} proposed to train entity linking module through a variational learning algorithm which jointly models topic entity recognition and subsequent reasoning over KBs. 
They also applied the REINFORCE algorithm with variance reduction technique to make the system end-to-end trainable.

%% file: sec-other.tex
\section{PLM Applications on Complex KBQA}
\label{sec:PLM}
Unsupervised pre-training language models on large text corpora then fine-tuning pre-trained language models (PLMs) on downstream tasks has become a popular paradigm for natural language processing~\cite{Foundation-arxiv-2021}.
Furthermore, due to the powerful performance obtained from broad data at scale and capability to serve a wide range of downstream tasks, PLMs are recognized as ``foundation models''~\cite{Han-arxiv-2021} for many tasks, including complex KBQA task.
Therefore, some recent SP-based and IR-based methods have widely incorporated PLMs in their pipelines.

For SP-based methods, PLMs are always used to simultaneously optimize the trainable modules~(\ie question understanding, logical parsing, KB grounding), which facilitate the generation of executable programs~(\eg SPARQL) in a seq2seq framework.
With such a unified paradigm, transferable knowledge across tasks can be leveraged to mitigate the data sparsity issue in low-resource scenarios.
For IR-based methods, PLMs help with precise source construction and further enhance the unified reasoning ability. 
On one hand, PLMs provide powerful representation ability to retrieve semantically relevant information from KBs.
On the other hand, PLMs can help unify the representation of questions and KBs, which contributes to the reasoning capability.


\subsection{PLM for SP-based Methods}
Equipped with powerful PLMs, the logic form generation modules benefit from their strong generation and understanding capabilities obtained via unsupervised pre-training. 
Under a unified seq2seq generation framework, PLMs provide transferable knowledge to help effective model training with limited data.

\paratitle{PLM for enhancing the logic form generation.}
To get the executable programs~(\eg SPARQL), traditional SP-based methods parse the question to a logic form and instantiate it via KB grounding.
This process can be well formalized under \textit{knowledge-enhanced text generation}~\cite{Yu-arxiv-2020} framework (\ie from user requests to executable programs).
Therefore, some work~\cite{Das-EMNLP-21, Ye-ACL-22} leveraged the PLMs, which are typically neural encoder-decoder models, to directly generate the executable programs given the question and other related KB information.
To get the related KB information input, Das \etal~\cite{Das-EMNLP-21} retrieved similar cases from a cases-memory where each case is a pair of question and its gold executable program.
Ye \etal~\cite{Ye-ACL-22} directly retrieved the top-$k$ relevant logic forms from the candidates enumerated by searching over the KB based on predefined rules.
Substantial improvement on model performance has proved the effectiveness of such usages of PLMs.

\paratitle{{PLM for low-resource training.}} 
The robust and transferable natural language understanding capability obtained from PLMs empowers KBQA methods to conquer the unaffordable need for training data in low-resource scenarios.
In a recent study, Shi \etal~\cite{KQA-Pro} fine-tuned pre-trained sequence-to-sequence model on KQA Pro dataset~\cite{KQA-Pro} to generate SPARQLs and programs. 
While no external knowledge was incorporated to enhance the generation, the BART-based generator reached near-human performance and showed robustness to sparse training data.

Besides, similar to complex KBQA, there are a series of structure parsing tasks (\eg Text2SQL, tabular question answering, semantic parsing over database) can be formed as the \textit{knowledge-enhanced text generation} framework. 
Motivated by this, Xie \etal~\cite{UnifiedSKG-arxiv-2022} proposed structured knowledge grounding (SKG) to unify a series of structure parsing tasks and achieved (near) state-of-the-art performance with PLM model T5~\cite{Raffel-JMLR-2020} on 21 benchmarks.  
With this PLM-based general-purpose approach, the challenges of precise semantic parsing in low-resource tasks can be solved by knowledge sharing and cross-task generalization.


\subsection{PLM for IR-based Methods}
With the strong representation ability of PLMs, we can augment the retrieval of the question-specific graphs and relieve the incompleteness of KBs during retrieval source construction.
Besides, PLMs provide a unified way to model unstructured text and structured KB information in a unified semantic space, which in turn improves question-specific graph reasoning.

\paratitle{PLM for augmenting the source construction.}
To cover the answers as complete as possible, traditional heuristic-based methods like personalized pagerank~\cite{PPR-Andersen-2006} would recall a large and noisy question-specific graph~\cite{Zhang-arxiv-2022}, which hinders subsequent reasoning.
Therefore, Zhang \etal~\cite{Zhang-arxiv-2022} trained a path retriever based on PLMs to retrieve the hop-wise question-related relations.
At each step, the retriever ranks top-$k$ relations based on the question and the selected relations of the previous step. This method successfully filters noisy graph context out and keeps high recall of expected reasoning paths for answers.

Besides building up a precise source construction module, PLMs also provide the potential to mitigate the incompleteness of KB. PLMs have shown their capabilities to answer "fill-in-the-blank" cloze statements~\cite{Petroni-EMNLP-2019,Bouraoui-AAAI-2020,Jiang-TACL-2020}, which indicates that they may learn relational knowledge from unsupervised pre-training. Such key findings indicate that PLMs have great potential in serving as knowledge sources for question answering, which may play a complementary role for source construction with incomplete KB.

\paratitle{PLM for precise and unified reasoning.}
Attracted by the powerful pre-trained language models, some researchers made adaptions to complex reasoning over graph structure for further involvement of PLM. 
While traditional reasoning over KB relies on the embeddings learned for entities and relations, such embeddings may fail to identify relevant parts of question answering context.
To filter out noisy graph context in the retrieved subgraph, Yasunaga\etal~\cite{Yasunaga-NAACL-2021} adopted PLM similarity scores to identify relevant knowledge given the question.
For further joint reasoning of question answering context (\ie question-answer sequence) and knowledge graph, the node representations in the retrieved subgraph were initialized with PLM encoding of the concatenated sequence of question, answer, and node surface name.
With the augmentation of PLMs, the GNN model gets substantial performance improvement~\cite{Yasunaga-NAACL-2021}.

%% file: sec-eval_resource.tex
\section{Evaluation and Resource}
\label{sec:evaluation_resource}
In this section, we first introduce the evaluation protocol of KBQA systems.
And then, we summarize some popular benchmarks for KBQA.
At last, for tracking the research progress conveniently, we make a leaderboard for these benchmark datasets, which contains the evaluation results and  resource links of the corresponding publications.
We also attach a companion page~\footnote{https://github.com/RUCAIBox/Awesome-KBQA} for comprehensive collection of the relevant publications, open-source codes, resources, and tools for KBQA.

\begin{table*}[!htbp]
\addtolength{\tabcolsep}{-3.0pt} 
\centering
	\caption{Several KBQA benchmark datasets involving complex questions. ``\textbf{LF}" denotes whether the dataset provides \textbf{L}ogic \textbf{F}orms like SPARQL, ``\textbf{CO}" denotes whether the dataset contains questions with \textbf{CO}nstraints, ``\textbf{NL}" represents whether the dataset incorporates crowd workers to rewrite questions in \textbf{N}atural \textbf{L}anguage and ``\textbf{NU}" denotes whether the dataset contains the questions which require \textbf{NU}merical operations. Typically, SP-based methods adopt $F_1$ score as evaluation metric, while IR-based methods adopt $Hits@1$ (accuracy) as evaluation metric. The symbol of $\triangle$ and $\heartsuit$ indicates evaluation metric of $Hits@1$ (accuracy) and $F_1$ score respectively.
	}
	\label{tab:datasets}%
	\begin{scriptsize}
	\begin{tabular}{c | c c | c c c c | c c c | c c c }
		\hline
		\multirow{2}{*}{Datasets}&	\multirow{2}{*}{KB}&	\multirow{2}{*}{Size}&	\multirow{2}{*}{LF}&	\multirow{2}{*}{CO} & \multirow{2}{*}{NL}& \multirow{2}{*}{NU} & \multicolumn{3}{c|}{SP-based} & \multicolumn{3}{c}{IR-based} \\
		\cline{8-13}
			&	&	&	&	& &	& Top-1 & Top-2 & Top-3 & Top-1 & Top-2 &  Top-3  \\
		\hline
		\hline
		WebQuestions~\cite{Berant-EMNLP-2013}&	Freebase&	5,810&	No&	Yes&	No&	Yes & 
		$62.9^{\heartsuit}$~\cite{Zheng-VLDB-2018} & $54.8^{\heartsuit}$~\cite{Qiu-CIKM-2020} & $54.6^{\heartsuit}$~\cite{Xu-NAACL-2019} & $48.6^{\triangle}$~\cite{Xu-NAACL-2019} & - & - \\
		ComplexQuestions~\cite{Bao-COLING-2016}&	Freebase&	2,100&	No&	Yes&	No&	Yes & 
		$71.0^{\heartsuit}$~\cite{Zheng-VLDB-2018} & $54.3^{\heartsuit}$~\cite{Hu-EMNLP-18} & $45.0^{\heartsuit}$~\cite{Qiu-CIKM-2020} & - & - & -\\
		WebQuestionsSP~\cite{Yih-ACL-2016}&	Freebase&	4,737&	Yes&	Yes&	Yes&	Yes & 
		$76.5^{\heartsuit}$~\cite{Cao-ACL-2022} & 
		$75.0^{\heartsuit}$~\cite{Zhang-KBS-2022} & 
		$74.0^{\heartsuit}$~\cite{Lan-ACL-2020} &
		$74.3^{\triangle}$~\cite{He-WSDM-2021} & $71.4^{\triangle}$~\cite{Shi-EMNLP-2021} & $69.5^{\triangle}$~\cite{Zhang-arxiv-2022}\\
        ComplexWebQuestions\cite{CWQ-NAACL-2018}& Freebase& 34,689& Yes& Yes& Yes& Yes & 
        $70.4^{\triangle}$~\cite{Das-EMNLP-21} & $44.1^{\triangle}$~\cite{Lan-ACL-2021} & $39.4^{\triangle}$~\cite{Lan-IJCAI-2019} & $53.9^{\triangle}$~\cite{He-WSDM-2021} &$45.9^{\triangle}$~\cite{Sun-EMNLP-2019}&-\\
		\hline
		QALD series~\cite{QALD}&	DBpedia&	-&	Yes&	Yes&	Yes&	Yes & - & - & - & - & - & - \\
        LC-QuAD~\cite{LC-QuAD-ISWC-2017}&	DBpedia&	5,000&	Yes&	Yes&	Yes&	Yes & 
        $75.0^{\heartsuit}$~\cite{Zafar-ESWC-2018} & $74.8^{\heartsuit}$~\cite{Chen-IJCAI-2020} & $71.8^{\heartsuit}$~\cite{Zheng-CIKM-2020} & $33.0^{\heartsuit}$~\cite{Vakulenko-CIKM-2019} & - & -\\
		\multirow{2}{*}{LC-QuAD 2.0~\cite{LC-QuAD-2.0-ISWC-2019}}&	\multirow{2}{*}{\tabincell{c}{DBpedia \&\\Wikidata}}&	\multirow{2}{*}{30,000}&	\multirow{2}{*}{Yes}&	\multirow{2}{*}{Yes}&	\multirow{2}{*}{Yes}&	\multirow{2}{*}{Yes}&
		\multirow{2}{*}{$59.3^{\heartsuit}$~\cite{Zou-arxiv-2021}} &
		\multirow{2}{*}{$52.6^{\heartsuit}$~\cite{Lan-ACL-2020}} &
		\multirow{2}{*}{$44.9^{\heartsuit}$~\cite{Chen-IJCAI-2020}} & \multirow{2}{*}{-}&\multirow{2}{*}{-}&\multirow{2}{*}{-} \\
		&	&	&	&	&	& & & & & & & \\
		\hline
		MetaQA	Vanilla~\cite{Zhang-AAAI-2018}&	WikiMovies&	400k&	No&	No&	No&	No & 
		$99.6^{\triangle}$~\cite{Lan-ICDM-2019} & - & - & $100.0^{\triangle}$~\cite{Shi-EMNLP-2021} & $99.3^{\triangle}$~\cite{Cai-ACL-2021} & $98.9^{\triangle}$~\cite{He-WSDM-2021}\\
        CFQ~\cite{CFQ-ICLR-2020}& Freebase&	239,357&	Yes&	Yes&	No&	No & 
        $67.3^{\triangle}$~\cite{Guo-NIPS-2020} & $18.9^{\triangle}$~\cite{CFQ-ICLR-2020} & - & - & - & -\\
		GrailQA~\cite{GrailQA-2020}&	Freebase&	64,331&	Yes&	Yes&	Yes&	Yes & 
		$74.4^{\heartsuit}$~\cite{Xiong-ACL-2022} & $65.3^{\heartsuit}$~\cite{Chen-ACL-2021} & $58.0^{\heartsuit}$~\cite{GrailQA-2020} & - & - & - \\
		KQA Pro~\cite{KQA-Pro}&	Wikidata&	117,970&	Yes&	Yes&	Yes&	Yes & 
		$89.7^{\heartsuit}$~\cite{KQA-Pro}  & - & - & - & - & - \\
		\hline
	\end{tabular}%
	\end{scriptsize}
\end{table*}%

\subsection{Evaluation Protocol}
In order to comprehensively evaluate KBQA systems, effective measurements from multiple aspects should be taken into consideration. 
Considering the goals to achieve, we categorize the measurement into three aspects: reliability, robustness, and system-user interaction~\cite{Diefenbach-KIS-2018}. 

\paratitle{Reliability}: 
For each question, there is an answer set (one or multiple elements) as the ground truth. 
The KBQA system usually predicts entities with the top confidence score to form the answer set. 
If an answer predicted by the KBQA system exists in the answer set, it is a correct prediction. 
In previous studies~\cite{Yih-ACL-2015,Liang-ACL-17,Abujabal-WWW-17}, there are some classical evaluation metrics such as Precision, Recall, $\text{F}_1$, and Hits@1.
For a question $q$, its Precision indicates the ratio of the correct predictions over all the predicted answers. 
It is formally defined as:
\begin{align*}
    \text{Precision} = \frac{|\mathcal{A}_q \cap \Tilde{\mathcal{A}}_q|}{|\Tilde{\mathcal{A}}_q|},
\end{align*}
where $\Tilde{\mathcal{A}}_q$ is the predicted answers, and  $\mathcal{A}_q$ is the ground truth.
Recall is the ratio of the correct predictions over all the ground truth.
It is computed as:
\begin{align*}
    \text{Recall} = \frac{|\mathcal{A}_q \cap \Tilde{\mathcal{A}}_q|}{|\mathcal{A}_q|}.
\end{align*}
Ideally, we expect that the KBQA system has a higher Precision and Recall simultaneously. 
Thus $\text{F}_1$ score is most commonly used to give a comprehensive evaluation:
\begin{align*}
    \text{F}_1 = \frac{2 * \text{Precision} * \text{Recall}}{\text{Precision} + \text{Recall}}.
\end{align*}
Some other methods~\cite{KVMem-EMNLP-2016,Sun-EMNLP-2018,Xiong-ACL-2019,He-WSDM-2021} use Hits@1 to assess the fraction that the correct prediction rank higher than other entities. It is computed as:
\begin{align*}
    \text{Hits@}1 = \mathbb{I}(\Tilde{a}_q\in \mathcal{A}_q),
\end{align*}
where $\Tilde{a}_q$ is the top $1$ prediction in $\Tilde{\mathcal{A}}_q$.


\paratitle{Robustness}: 
Practical KBQA models are supposed to generalize to out-of-distribution questions at test time~\cite{GrailQA-2020}. 
However, current KBQA datasets are mostly generated based on templates and lack of diversity~\cite{Diefenbach-KIS-2018}.
And, the scale of training datasets is limited by the expensive labeling cost. 
Furthermore, the training data for KBQA system may hardly cover all possible user queries due to broad coverage and combinatorial explosion of queries.
To promote the robustness of KBQA models, Gu \etal~\cite{GrailQA-2020} proposed three levels of generalization (\ie \textit{i.i.d.}, \textit{compositional}, and \textit{zero-shot}) and released a large-scale KBQA dataset GrailQA to support further research. 
At a basic level, KBQA models are assumed to be trained and tested with questions drawn from the same distribution, which is what most existing studies focus on. 
In addition to that, robust KBQA models can generalize to novel compositions of seen schema items (\eg relations and entity types). 
To achieve better generalization and serve users, robust KBQA models are supposed to handle questions whose schema items or domains are not covered in the training stage.
\paratitle{System-user Interaction}: 
While most of the current studies pay much attention to offline evaluation, the interaction between users and KBQA systems is neglected. 
On one hand, in the search scenarios, a user-friendly interface and acceptable response time should be taken into consideration.
To evaluate this, the feedback of users should be collected and the efficiency of the system should be judged.
On the other hand, users' search intents may be easily misunderstood by systems if only a single round service is provided.
Therefore, it is important to evaluate the interaction capability of a KBQA system.
For example, to check whether they could ask clarification questions to disambiguate users' queries and whether they could respond to the error reported from the users~\cite{Abujabal-WWW-2018,Yao-EMNLP-2020}. 
So far, there is a lack of quantitative measurement of system-user interaction capability of the system, but human evaluation can be regarded as an efficient and comprehensive way.

\subsection{Datasets And Leaderboard}
\label{sec:data}
\paratitle{Datasets.}
Over the decades, much effort has been devoted to constructing datasets for complex KBQA. 
We list the representative complex KBQA datasets for multiple popular KBs~(\eg Freebase, DBpedia, Wikidata, and WikiMovies) in Table~\ref{tab:datasets}.
In order to serve realistic applications, these datasets typically contain questions which require multiple KB facts to reason. Moreover, they might include numerical operations (\eg counting and ranking operations for comparative and superlative questions, respectively) and constraints (\eg entity and temporal keywords), which further increase the difficulty in reasoning the answers from KBs.

Overall, these datasets are constructed with the following steps: Given a topic entity in a KB as a question subject, simple questions are first created with diverse templates. Based on simple questions and neighborhood of topic entities in a KB, complex questions are further generated with predefined composition templates, and another work~\cite{KQA-Pro} also generates executable logic forms with templates. 
Meanwhile, answers are extracted with corresponding rules. 
In some cases, crowd workers are hired to paraphrase the canonical questions, and refine the generated logic forms, making the question expressions more diverse and fluent.

\paratitle{Leaderboard.}
In order to show the latest research progress in these KBQA benchmark datasets, we offer a leaderboard including the top-3 KBQA systems with respect to IR-based methods and SP-based methods.
To give a fair comparison, the  results are selected following three principles: 1) 
If one dataset has an official leaderboard, we only refer to the public results listed on the leaderboard. 2) otherwise, we select the top-3 results from the published papers accepted formally by conferences or journals before March 2022. 3) besides, we keep the experimental setup of each dataset consistent with the other datasets.
Exceptionally, we do not report the results on the QALD series for easy display because they have multiple different versions, and we only report the 3-hop split of MetaQA Vanilla because it is more challenging than 1-hop and 2-hop splits.
For LC-QuAD~2.0, we select the results reported in Zou~\etal~\cite{Zou-arxiv-2021}, where they reproduced the top 2 and top 3 results using the official public codes of respective methods and obtained the top 1 result by their proposed model. 
We leave the slots blank if there is no sufficient result according to the above principles. 
For more comprehensive evaluation of KBQA methods on all benchmarks, please refer to our companion page.

\paratitle{Analysis and discussions.}
Based on Table~\ref{tab:datasets}, we have following observations: (1) Both SP-based and IR-based methods are developed to handle complex KBQA challenges and there is no absolute agreement on which category is better. (2) While SP-based methods cover most benchmarks, IR-based methods focus on benchmarks which are mainly composed of multi-hop questions. The reason why SP-based methods are more commonly used in answering complex questions may be that SP-based methods generate flexible and expressive logic forms which are capable of covering all types of questions (\eg boolean, comparative).
(3) We also observe for each category, the methods achieving outstanding performance are usually equipped with advanced techniques. The SP-based methods on the leaderboard leverage powerful question encoders (\eg PLMs) to help understand the questions and expressive logic forms to help parse complex queries.
For IR-based methods, most SOTA methods adopt the step-wise dynamic instruction in question representation module and conduct multi-step reasoning with relational path modeling or GNN-based reasoning.



%% file: sec-con.tex
\section{Recent Trends}
\label{sec:direction}

In this section, we discuss several promising future directions for complex KBQA task:

\ignore{
\paratitle{Improve reasoning pipeline.}~\yscomment{This direction is too general} 
As we can see that, both semantic parsing-based methods and information retrieval-based methods seek for better training targets, reasoning interpretability, inference accuracy via different techniques, such as incorporating extra information, designing efficient operations, constraining intermediate reasoning process. 
Following these motivations, improving the whole complex KBQA pipeline (\eg entity linking, target setting) will keep playing an important role in this research topic.
}

\paratitle{Evolutionary KBQA systems.} 
Existing KBQA systems are usually trained offline with specific datasets and then deployed online to handle user queries. 
However, most of existing KBQA systems neglect to learn from failure cases or unseen question templates received after deployment. 
At the same time, most of existing KBQA systems fail to catch up with the rapid growth of world knowledge and answer new questions. 
Therefore, a practical KBQA system is imperative to get performance  improvement over time after online deployment.
Online user interaction may provide deployed KBQA systems an opportunity to get further improvement. 
Based on this motivation, some work leverages user interaction to rectify answers generated by the KBQA system and further improve itself. With user feedback, Abujabal~\etal~\cite{Abujabal-WWW-2018} presented a continuous learning framework to learn new templates that capture previously unseen syntactic structures. 
\ignore{
Encountering new questions after online deployment, they leveraged user feedback to identify valid logic forms among candidates retrieved from similar questions. With the valid logic forms, new templates were  generated to handle syntactic variation and keep KBQA system evolving.} 
Besides increasing the model's template bank, user feedback can also be leveraged to clarify ambiguous questions (\eg ambiguous phrases or ambiguous entities)~\cite{Zheng-IS-2019}. 
Above methods provide an initial exploration to construct evolutionary KBQA systems with user feedback. 
Such approaches are effective and practical (\ie acceptable user cognitive burden and running cost), which may serve the industrial needs. 
Due to the wide applications of KBQA systems, more work and designs of user interaction with KBQA systems are in urgent need.


\paratitle{Robust KBQA systems.} 
Existing studies on KBQA have conducted with the ideal hypothesis, where training data is sufficient and its distribution is identical with test set. 
However, this may not be desirable in practice due to data insufficiency and potential data distributional biases. 
To train robust KBQA systems in low-resource scenarios, meta-learning techniques~\cite{Hua-EMNLP-2020} and knowledge transfer from high-resource scenarios~\cite{Zhou-NAACL-2021} have been explored. We also highlighted the potential impact of PLMs in low-resource training and cross-task generalization (see Sec~\ref{sec:PLM}). As the manual annotations for KBQA systems are expensive and labor-intensive, there is a need for more studies about training robust KBQA systems in low-resource scenarios.
Meanwhile, although existing methods usually hold the i.i.d. assumption, they may easily fail to deal with out-of-distribution (OOD) issue~\cite{Gardner-arXiv-2020,Kaushik-ICLR-2020,Oren-EMNLP-2020} on KBQA.
With a systematic evaluation of GrailQA~\cite{GrailQA-2020} dataset, Gu \etal~\cite{GrailQA-2020} pointed out that existing baseline methods are vulnerable to compositional challenges.
To promote higher level of robustness, researchers may gain more insights with addressing the three levels of generalization (\ie \textit{i.i.d.}, \textit{compositional}, and \textit{zero-shot}) proposed by Gu \etal~\cite{GrailQA-2020}.
There is few work investigating robustness on complex KBQA task. 
It is still an open question of building robust KBQA systems with stronger generalization capability.

\ignore{
While existing methods have achieved promising results on benchmark datasets, they may easily fail to deal with a similar but unseen question. 
To address such issue, the models are supposed to be able to handle out-of-distribution questions~\cite{GrailQA-2020} and keep awareness of the reasoning process. 
It is interesting to analyze the compositional generalization capabilities of the complex KBQA methods and propose a robust method with good interpretability.
~\glcomment{Useful citation for robust and interpretable models: robust adversarial training~\cite{Zhang-Access-2020}, interpretable system~\cite{Abujabal-EMNLP-2017}}
}


\paratitle{Conversational KBQA systems.} 
Recent decades have seen the rapid development of AI-driven applications (\eg search engines and personal assistants) which are supposed to answer factoid questions.
As users typically ask follow-up questions to explore a topic, deployed models are supposed to handle KBQA task in a conversational manner.
In initial explorations of conversational KBQA, several pieces of work~\cite{Guo-NIPS-2018,Christmann-CIKM-2019,Plepi-ESWC-2021,kaiser:sigir2021} focused on ambiguity and difficulties brought by coreference and ellipsis phenomena. To track the focus of conversational KBQA, Lan \etal~\cite{Lan-ACL-2021} proposed to model the flow of the focus via an entity transition graph. 
For a comprehensive understanding of conversation context, Plepi \etal~\cite{Plepi-ESWC-2021,kacupaj:eacl2021} leveraged Transformer~\cite{Vaswani-NIPS-2017} architecture taking as input of the previous turn of conversation history. While these studies addressed some challenges for conversational KBQA, it is still far from achieving human-level performance. More critical challenges should be identified and solved in the following research.
Up to now, conversational KBQA is quite a new and challenging task, it may play an important role in future search engines and intelligent personal assistants.

\paratitle{Neural symbolic KBQA systems.} 
While some recent work\cite{Liang-ACL-17,Sun-EMNLP-2018} has proposed to equip KBQA systems with neural symbolic reasoning~(NSR) techniques, the promising potential of such powerful paradigm has not been explored thoroughly. 
For example, while neural networks have been proved to be effective in conduct multi-hop reasoning on KBs\cite{Sun-EMNLP-2018,Yasunaga-NAACL-2021,He-WSDM-2021}, such neural modules can not explicitly consider logical operations (\eg numerical, boolean). 
To mitigate such drawbacks while keeping power of neural networks in reasoning, we can introduce a symbolic module coupled with existing neural reasoning modules~\cite{Lamb-IJCAI-2020}. Several practices in neural programming\cite{Arabshahi-ICLR-2018,Lample-ICLR-2020} have demonstrated this can be effective in empowering blackbox neural networks with mathematical and logical reasoning capabilities. 
In general, researchers appreciate the interpretability of SP-based methods (\ie generating logic forms according to grammar rules) and the powerfulness of IR-based methods (\ie precise reasoning on subgraph with neural networks). 
As discussed in section~\ref{sec-NSR}, both SP-based and IR-based methods can be unified paradigm --- neural symbolic reasoning. 
Thus, NSR provides a potential way to unify the two categories of methods and gather their advantages, which deserves further research. 

\paratitle{More general knowledge bases.} 
Due to KB incompleteness, researchers incorporated extra information (such as text~\cite{Xie-AAAI-2016}, images~\cite{Xie-IJCAI-2017}, and human interactions~\cite{He-WWW-2020}) to complement the knowledge bases, which would further address the information need of complex KBQA task.
As text corpus is rich in semantics and easy to collect, researchers are fascinated by the idea of extracting knowledge from text corpus and answering questions with extracted knowledge. 
Researchers have explored various forms of knowledge obtained from text corpus, such as traditional relational triplet~\cite{Lu-SIGIR-2019}, virtual knowledge base (VKB)~\cite{Dhingra-ICLR-2020} which is stored as key-value memory, and PLMs as implicit knowledge base~\cite{Petroni-EMNLP-2019}. 
With these elaborate designs, more flexible and complementary knowledge can be obtained to solve complex KBQA tasks.
Recently, a neglectable trend is to unify similar tasks with general architecture and achieve cross-task knolwedge transfer~\cite{UnifiedSKG-arxiv-2022}. In the future, more related tasks may be explored with a general definition of KBs, such as synthetic, multilingual, and multi-modal KBs. 






\section{Conclusion}
\label{sec:con}
\ignore{
This paper attempted to provide an overview of recent developments on complex KBQA. 
Obviously, it cannot cover all the literature on complex KBQA, and we focused on a representative subset of the typical challenges and advanced solutions.
}
This survey attempted to provide an overview of typical challenges and corresponding solutions on complex KBQA. 
Particularly, task-related preliminary knowledge and traditional methods were first introduced.
Then, we summarized the widely employed semantic parsing-based methods and information retrieval-based methods. 
We specified the challenges for these two categories of methods based on their working mechanism, and explicated the proposed solutions.
Along with the taxonomy, we provide technical summaries to shed light on the applied advanced techniques for these two categories.
Most of existing complex KBQA methods are generally summarized into these two categories.
Please be aware that there are some other methods like~\cite{CWQ-NAACL-2018}, which focus on question decomposition instead of KB based reasoning or logic form generation.
In the last section, we investigated several research trends related to complex KBQA task and emphasized many challenges are still open and under-explored. 
We believe that complex KBQA will continue to be an active and promising research area with wide applications, such as natural language understanding, compositional generalization, multi-hop reasoning. 
We hope this survey will give a comprehensive picture of cutting-edge methods for complex KBQA and encourage further contributions in this field.

\ignore{
\glcomment{Here, we show the setup for future directions, but this list should be deleted later.}
\begin{itemize}
	\item Develop Expressive Targets for Parsing, Narrow Down the Search Space $\rightarrow$ Improved reasoning pipeline.
	\item Train under Weak Supervision, Promote Interpretability of Reasoning Process, Reason under Weak Supervision. $\rightarrow$ Robust and interpretable model.
	\item Augment Complex Parsing with Structural Properties. $\rightarrow$ Understand complex question with syntactic knowledge.
	\item Supplement Incomplete Knowledge Bases. $\rightarrow$ More general grounded knowledge base.
\end{itemize}
}

%% file: main.bbl
\begin{thebibliography}{100}
\providecommand{\url}[1]{#1}
\csname url@samestyle\endcsname
\providecommand{\newblock}{\relax}
\providecommand{\bibinfo}[2]{#2}
\providecommand{\BIBentrySTDinterwordspacing}{\spaceskip=0pt\relax}
\providecommand{\BIBentryALTinterwordstretchfactor}{4}
\providecommand{\BIBentryALTinterwordspacing}{\spaceskip=\fontdimen2\font plus
\BIBentryALTinterwordstretchfactor\fontdimen3\font minus
  \fontdimen4\font\relax}
\providecommand{\BIBforeignlanguage}[2]{{%
\expandafter\ifx\csname l@#1\endcsname\relax
\typeout{** WARNING: IEEEtran.bst: No hyphenation pattern has been}%
\typeout{** loaded for the language `#1'. Using the pattern for}%
\typeout{** the default language instead.}%
\else
\language=\csname l@#1\endcsname
\fi
#2}}
\providecommand{\BIBdecl}{\relax}
\BIBdecl

\bibitem{Bollacker-SIGMOD-2008}
K.~Bollacker, C.~Evans, P.~Paritosh, T.~Sturge, and J.~Taylor, ``Freebase: A
  collaboratively created graph database for structuring human knowledge,'' in
  \emph{{SIGMOD}}, 2008.

\bibitem{dbpedia-2015}
J.~Lehmann, R.~Isele, M.~Jakob, A.~Jentzsch, D.~Kontokostas, P.~N. Mendes,
  S.~Hellmann, M.~Morsey, P.~Van~Kleef, S.~Auer \emph{et~al.}, ``Dbpedia--a
  large-scale, multilingual knowledge base extracted from wikipedia,''
  \emph{Semantic Web}, 2015.

\bibitem{Thomas-WWW-2016}
T.~P. Tanon, D.~Vrandecic, S.~Schaffert, T.~Steiner, and L.~Pintscher, ``From
  freebase to wikidata: The great migration,'' in \emph{{WWW}}, 2016.

\bibitem{Suchanek-WWW-2007}
F.~M. Suchanek, G.~Kasneci, and G.~Weikum, ``Yago: A core of semantic
  knowledge,'' in \emph{{WWW}}, 2007.

\bibitem{Zhou-IJCAI-2018}
H.~Zhou, T.~Young, M.~Huang, H.~Zhao, J.~Xu, and X.~Zhu, ``Commonsense
  knowledge aware conversation generation with graph attention,'' in
  \emph{{IJCAI}}, 2018.

\bibitem{Bordes-ArXiv-2015}
A.~Bordes, N.~Usunier, S.~Chopra, and J.~Weston, ``Large-scale simple question
  answering with memory networks,'' \emph{CoRR}, 2015.

\bibitem{Dong-ACL-2015}
L.~Dong, F.~Wei, M.~Zhou, and K.~Xu, ``Question answering over freebase with
  multi-column convolutional neural networks,'' in \emph{{ACL}}, 2015.

\bibitem{Yu-ACL-2017}
M.~Yu, W.~Yin, K.~S. Hasan, C.~N. dos Santos, B.~Xiang, and B.~Zhou, ``Improved
  neural relation detection for knowledge base question answering,'' in
  \emph{{ACL}}, 2017.

\bibitem{Petrochuk-EMNLP-2018}
M.~Petrochuk and L.~Zettlemoyer, ``Simplequestions nearly solved: {A} new
  upperbound and baseline approach,'' in \emph{{EMNLP}}, 2018.

\bibitem{Hu-TKDE-2018}
S.~Hu, L.~Zou, J.~X. Yu, H.~Wang, and D.~Zhao, ``Answering natural language
  questions by subgraph matching over knowledge graphs,'' \emph{TKDE}, 2018.

\bibitem{Hu-EMNLP-18}
S.~Hu, L.~Zou, and X.~Zhang, ``A state-transition framework to answer complex
  questions over knowledge base,'' in \emph{{EMNLP}}, 2018.

\bibitem{Luo-EMNLP-2018}
K.~Luo, F.~Lin, X.~Luo, and K.~Q. Zhu, ``Knowledge base question answering via
  encoding of complex query graphs,'' in \emph{{EMNLP}}, 2018.

\bibitem{GrailQA-2020}
Y.~Gu, S.~Kase, M.~Vanni, B.~M. Sadler, P.~Liang, X.~Yan, and Y.~Su, ``Beyond
  {I.I.D.:} three levels of generalization for question answering on knowledge
  bases,'' in \emph{{WWW}}, 2021.

\bibitem{Wu-CCK-2019}
P.~Wu, X.~Zhang, and Z.~Feng, ``A survey of question answering over knowledge
  base,'' in \emph{{CCKS}}, 2019.

\bibitem{Chakraborty-arXiv-2019}
N.~Chakraborty, D.~Lukovnikov, G.~Maheshwari, P.~Trivedi, J.~Lehmann, and
  A.~Fischer, ``Introduction to neural network based approaches for question
  answering over knowledge graphs,'' \emph{CoRR}, 2019.

\bibitem{gu:akbc2022}
Y.~Gu, V.~Pahuja, G.~Cheng, and Y.~Su, ``Knowledge base question answering: A
  semantic parsing perspective,'' in \emph{{AKBC}}, 2022.

\bibitem{Fu-Arxiv-2020}
B.~Fu, Y.~Qiu, C.~Tang, Y.~Li, H.~Yu, and J.~Sun, ``A survey on complex
  question answering over knowledge base: Recent advances and challenges,''
  \emph{CoRR}, 2020.

\bibitem{Lan-IJCAI-2021}
Y.~Lan, G.~He, J.~Jiang, J.~Jiang, W.~X. Zhao, and J.-R. Wen, ``A survey on
  complex knowledge base question answering: Methods, challenges and
  solutions,'' in \emph{{IJCAI}}, 2021.

\bibitem{freebase}
Google, ``Freebase data dumps,''
  \url{https://developers.google.com/freebase/data}, 2016.

\bibitem{Cyganiak-WWW-2014}
R.~Cyganiak, D.~Wood, and M.~Lanthaler, ``Rdf 1.1 concepts and abstract
  syntax,'' in \emph{{WWW}}, 2014.

\bibitem{yago-2016}
T.~Rebele, F.~M. Suchanek, J.~Hoffart, J.~Biega, E.~Kuzey, and G.~Weikum,
  ``{YAGO:} {A} multilingual knowledge base from wikipedia, wordnet, and
  geonames,'' in \emph{{ISWC}}, 2016.

\bibitem{wordnet-2015}
G.~A. Miller, ``Wordnet: {A} lexical database for english,'' \emph{Commun.
  {ACM}}, 1995.

\bibitem{farber-2015}
M.~F{\"a}rber, B.~Ell, C.~Menne, and A.~Rettinger, ``A comparative survey of
  dbpedia, freebase, opencyc, wikidata, and yago,'' \emph{Semantic Web}, 2015.

\bibitem{Yang-ACL-2015}
Y.~Yang and M.~Chang, ``{S}-{MART}: Novel tree-based structured learning
  algorithms applied to tweet entity linking,'' in \emph{{ACL}}, 2015.

\bibitem{Daiber-IS-2013}
J.~Daiber, M.~Jakob, C.~Hokamp, and P.~N. Mendes, ``Improving efficiency and
  accuracy in multilingual entity extraction,'' in \emph{I-SEMANTICS}, 2013.

\bibitem{Yosef-VLDB-2011}
M.~A. Yosef, J.~Hoffart, I.~Bordino, M.~Spaniol, and G.~Weikum, ``{AIDA:} an
  online tool for accurate disambiguation of named entities in text and
  tables,'' \emph{{VLDB}}, 2011.

\bibitem{Cai-ACL-2013}
Q.~Cai and A.~Yates, ``Large-scale semantic parsing via schema matching and
  lexicon extension,'' in \emph{{ACL}}, 2013.

\bibitem{Kwiatkowski-EMNLP-2013}
T.~Kwiatkowski, E.~Choi, Y.~Artzi, and L.~Zettlemoyer, ``Scaling semantic
  parsers with on-the-fly ontology matching,'' in \emph{{EMNLP}}, 2013.

\bibitem{Reddy-TACL-2014}
S.~Reddy, M.~Lapata, and M.~Steedman, ``Large-scale semantic parsing without
  question-answer pairs,'' \emph{{TACL}}, 2014.

\bibitem{Unger-WWW-2012}
C.~Unger, L.~B{\"{u}}hmann, J.~Lehmann, A.~N. Ngomo, D.~Gerber, and P.~Cimiano,
  ``Template-based question answering over {RDF} data,'' in \emph{{WWW}}, 2012.

\bibitem{Berant-EMNLP-2013}
J.~Berant, A.~Chou, R.~Frostig, and P.~Liang, ``Semantic parsing on freebase
  from question-answer pairs,'' in \emph{{EMNLP}}, 2013.

\bibitem{Yih-ACL-2014}
W.~Yih, X.~He, and C.~Meek, ``Semantic parsing for single-relation question
  answering,'' in \emph{{ACL}}, 2014.

\bibitem{Yao-ACL-2014}
X.~Yao and B.~V. Durme, ``Information extraction over structured data: Question
  answering with {F}reebase,'' in \emph{{ACL}}, 2014.

\bibitem{Yih-ACL-2016}
W.~Yih, M.~Richardson, C.~Meek, M.~Chang, and J.~Suh, ``The value of semantic
  parse labeling for knowledge base question answering,'' in \emph{{ACL}},
  2016.

\bibitem{Yih-ACL-2015}
W.~Yih, M.~Chang, X.~He, and J.~Gao, ``Semantic parsing via staged query graph
  generation: Question answering with knowledge base,'' in \emph{{ACL}}, 2015.

\bibitem{Reddy-TACL-2016}
S.~Reddy, O.~T{\"a}ckstr{\"o}m, M.~Collins, T.~Kwiatkowski, D.~Das,
  M.~Steedman, and M.~Lapata, ``Transforming dependency structures to logical
  forms for semantic parsing,'' \emph{{TACL}}, 2016.

\bibitem{Yin-COLING-2016}
W.~Yin, M.~Yu, B.~Xiang, B.~Zhou, and H.~Sch{\"{u}}tze, ``Simple question
  answering by attentive convolutional neural network,'' in \emph{{COLING}},
  2016.

\bibitem{Lan-TASLP-2019}
Y.~Lan, S.~Wang, and J.~Jiang, ``Knowledge base question answering with a
  matching-aggregation model and question-specific contextual relations,''
  \emph{{TASLP}}, 2019.

\bibitem{Andreas-NAACL-2016}
J.~Andreas, M.~Rohrbach, T.~Darrell, and D.~Klein, ``Learning to compose neural
  networks for question answering,'' in \emph{{NAACL-HLT}}, 2016.

\bibitem{Bordes-PKDD-2014}
A.~Bordes, J.~Weston, and N.~Usunier, ``Open question answering with weakly
  supervised embedding models,'' in \emph{Machine Learning and Knowledge
  Discovery in Databases}, 2014.

\bibitem{Bordes-EMNLP-2014}
A.~Bordes, S.~Chopra, and J.~Weston, ``Question answering with subgraph
  embeddings,'' in \emph{{EMNLP}}, 2014.

\bibitem{Sukhbaatar-NIPS-2015}
S.~Sukhbaatar, A.~Szlam, J.~Weston, and R.~Fergus, ``End-to-end memory
  networks,'' in \emph{{NeurIPS}}, 2015.

\bibitem{KVMem-EMNLP-2016}
A.~H. Miller, A.~Fisch, J.~Dodge, A.~Karimi, A.~Bordes, and J.~Weston,
  ``Key-value memory networks for directly reading documents,'' in
  \emph{{EMNLP}}, 2016.

\bibitem{Jain-NAACL-2016}
S.~Jain, ``Question answering over knowledge base using factual memory
  networks,'' in \emph{{NAACL-HLT}}, 2016.

\bibitem{Hao-ACL-2017}
Y.~Hao, Y.~Zhang, K.~Liu, S.~He, Z.~Liu, H.~Wu, and J.~Zhao, ``An end-to-end
  model for question answering over knowledge base with cross-attention
  combining global knowledge,'' in \emph{{ACL}}, 2017.

\bibitem{Chen-NAACL-2019}
Z.-Y. Chen, C.-H. Chang, Y.-P. Chen, J.~Nayak, and L.-W. Ku, ``{UH}op: An
  unrestricted-hop relation extraction framework for knowledge-based question
  answering,'' in \emph{{NAACL-HLT}}, 2019.

\bibitem{Hochreiter-NeuralComput-1997}
S.~Hochreiter and J.~Schmidhuber, ``Long short-term memory,'' \emph{Neural
  Comput.}, 1997.

\bibitem{Cho-SSST-2014}
K.~Cho, B.~van Merri{\"e}nboer, D.~Bahdanau, and Y.~Bengio, ``On the properties
  of neural machine translation: Encoder{--}decoder approaches,'' in
  \emph{Proceedings of {SSST}-8, Eighth Workshop on Syntax, Semantics and
  Structure in Statistical Translation}, 2014.

\bibitem{Liang-ACL-17}
C.~Liang, J.~Berant, Q.~Le, K.~D. Forbus, and N.~Lao, ``Neural symbolic
  machines: Learning semantic parsers on {F}reebase with weak supervision,'' in
  \emph{{ACL}}, 2017.

\bibitem{Sun-EMNLP-2018}
H.~Sun, B.~Dhingra, M.~Zaheer, K.~Mazaitis, R.~Salakhutdinov, and W.~W. Cohen,
  ``Open domain question answering using early fusion of knowledge bases and
  text,'' in \emph{{EMNLP}}, 2018.

\bibitem{Xiong-ACL-2019}
W.~Xiong, M.~Yu, S.~Chang, X.~Guo, and W.~Y. Wang, ``Improving question
  answering over incomplete kbs with knowledge-aware reader,'' in \emph{{ACL}},
  2019.

\bibitem{Diefenbach-KIS-2018}
D.~Diefenbach, V.~L{\'{o}}pez, K.~D. Singh, and P.~Maret, ``Core techniques of
  question answering systems over knowledge bases: a survey,'' \emph{Knowl.
  Inf. Syst.}, 2018.

\bibitem{Abujabal-WWW-17}
A.~Abujabal, M.~Yahya, M.~Riedewald, and G.~Weikum, ``Automated template
  generation for question answering over knowledge graphs,'' in \emph{{WWW}},
  2017.

\bibitem{Abujabal-WWW-2018}
A.~Abujabal, R.~S. Roy, M.~Yahya, and G.~Weikum, ``Never-ending learning for
  open-domain question answering over knowledge bases,'' in \emph{{WWW}}, 2018.

\bibitem{Kapanipathi-AAAI-2021}
P.~Kapanipathi, I.~Abdelaziz, S.~Ravishankar, S.~Roukos, A.~G. Gray, R.~F.
  Astudillo, M.~Chang, C.~Cornelio, S.~Dana, A.~Fokoue, D.~Garg, A.~Gliozzo,
  S.~Gurajada, H.~Karanam, N.~Khan, D.~Khandelwal, Y.~Lee, Y.~Li, F.~P.~S.
  Luus, N.~Makondo, N.~Mihindukulasooriya, T.~Naseem, S.~Neelam, L.~Popa, R.~G.
  Reddy, R.~Riegel, G.~Rossiello, U.~Sharma, G.~P.~S. Bhargav, and M.~Yu,
  ``Leveraging abstract meaning representation for knowledge base question
  answering,'' in \emph{{Findings of ACL}}, 2021.

\bibitem{Skeleton-AAAI-2020}
Y.~Sun, L.~Zhang, G.~Cheng, and Y.~Qu, ``{SPARQA:} skeleton-based semantic
  parsing for complex questions over knowledge bases,'' in \emph{{AAAI}}, 2020.

\bibitem{Zhu-Neuro-2020}
S.~Zhu, X.~Cheng, and S.~Su, ``Knowledge-based question answering by
  tree-to-sequence learning,'' \emph{Neurocomputing}, 2020.

\bibitem{Zafar-ESWC-2018}
H.~Zafar, G.~Napolitano, and J.~Lehmann, ``Formal query generation for question
  answering over knowledge bases,'' in \emph{{ESWC}}, 2018.

\bibitem{Chen-IJCAI-2020}
Y.~Chen, H.~Li, Y.~Hua, and G.~Qi, ``Formal query building with query structure
  prediction for complex question answering over knowledge base,'' in
  \emph{{IJCAI}}, 2020.

\bibitem{Bast-CIKM-15}
H.~Bast and E.~Haussmann, ``More accurate question answering on freebase,'' in
  \emph{{CIKM}}, 2015.

\bibitem{Jia-CIKM-2018}
Z.~Jia, A.~Abujabal, R.~Saha~Roy, J.~Str\"{o}tgen, and G.~Weikum, ``Tequila:
  Temporal question answering over knowledge bases,'' in \emph{{CIKM}}, 2018.

\bibitem{Zheng-VLDB-2018}
W.~Zheng, J.~X. Yu, L.~Zou, and H.~Cheng, ``Question answering over knowledge
  graphs: Question understanding via template decomposition,'' 2018.

\bibitem{Bhutani-CIKM-2019}
N.~Bhutani, X.~Zheng, and H.~V. Jagadish, ``Learning to answer complex
  questions over knowledge bases with query composition,'' in \emph{{CIKM}},
  2019.

\bibitem{Lan-ICDM-2019}
Y.~Lan, S.~Wang, and J.~Jiang, ``Multi-hop knowledge base question answering
  with an iterative sequence matching model,'' in \emph{{ICDM}}, 2019.

\bibitem{Lan-ACL-2020}
Y.~Lan and J.~Jiang, ``Query graph generation for answering multi-hop complex
  questions from knowledge bases,'' in \emph{{ACL}}, 2020.

\bibitem{Saha-TACL-2019}
A.~Saha, G.~A. Ansari, A.~Laddha, K.~Sankaranarayanan, and S.~Chakrabarti,
  ``Complex program induction for querying knowledge bases in the absence of
  gold programs,'' \emph{{TACL}}, 2019.

\bibitem{Qiu-CIKM-2020}
Y.~Qiu, K.~Zhang, Y.~Wang, X.~Jin, L.~Bai, S.~Guan, and X.~Cheng,
  ``Hierarchical query graph generation for complex question answering over
  knowledge graph,'' in \emph{{CIKM}}, 2020.

\bibitem{Qiu-WSDM-2020}
Y.~Qiu, Y.~Wang, X.~Jin, and K.~Zhang, ``Stepwise reasoning for multi-relation
  question answering over knowledge graph with weak supervision,'' in
  \emph{{WSDM}}, 2020.

\bibitem{Hua-JWS-2020}
Y.~Hua, Y.~Li, G.~Qi, W.~Wu, J.~Zhang, and D.~Qi, ``Less is more:
  Data-efficient complex question answering over knowledge bases,'' \emph{J.
  Web Semant.}, 2020.

\bibitem{gal:vldb2015}
L.~Gal\'{a}rraga, C.~Teflioudi, K.~Hose, and F.~M. Suchanek, ``Fast rule mining
  in ontological knowledge bases with amie++,'' \emph{The VLDB Journal},
  vol.~24, no.~6, p. 707–730, 2015.

\bibitem{lamb:ijcai2020}
L.~C. Lamb, A.~d. Garcez, M.~Gori, M.~O. Prates, P.~H. Avelar, and M.~Y. Vardi,
  ``Graph neural networks meet neural-symbolic computing: A survey and
  perspective,'' in \emph{{IJCAI}}, 2020.

\bibitem{Zhang-arxiv-2021}
J.~Zhang, B.~Chen, L.~Zhang, X.~Ke, and H.~Ding, ``Neural, symbolic and
  neural-symbolic reasoning on knowledge graphs,'' \emph{CoRR}, 2021.

\bibitem{gal:www2013}
L.~A. Gal\'{a}rraga, C.~Teflioudi, K.~Hose, and F.~Suchanek, ``Amie:
  Association rule mining under incomplete evidence in ontological knowledge
  bases,'' in \emph{Proceedings of the 22nd International Conference on World
  Wide Web}, ser. WWW '13, 2013, p. 413–422.

\bibitem{sun:aaai2020}
Y.~Sun, D.~Tang, N.~Duan, Y.~Gong, X.~Feng, B.~Qin, and D.~Jiang, ``Neural
  semantic parsing in low-resource settings with back-translation and
  meta-learning,'' in \emph{{AAAI}}, 2020.

\bibitem{Sun-EMNLP-2019}
H.~Sun, T.~Bedrax{-}Weiss, and W.~W. Cohen, ``Pullnet: Open domain question
  answering with iterative retrieval on knowledge bases and text,'' in
  \emph{{EMNLP}}, 2019.

\bibitem{Bao-COLING-2016}
J.~Bao, N.~Duan, Z.~Yan, M.~Zhou, and T.~Zhao, ``Constraint-based question
  answering with knowledge graph,'' in \emph{{COLING}}, 2016.

\bibitem{Maheshwari-ISWC-2019}
G.~Maheshwari, P.~Trivedi, D.~Lukovnikov, N.~Chakraborty, A.~Fischer, and
  J.~Lehmann, ``Learning to rank query graphs for complex question answering
  over knowledge graphs,'' in \emph{{ISWC}}, 2019.

\bibitem{Tai-ACL-2015}
K.~S. Tai, R.~Socher, and C.~D. Manning, ``Improved semantic representations
  from tree-structured long short-term memory networks,'' in \emph{{ACL}},
  2015.

\bibitem{CWQ-NAACL-2018}
A.~Talmor and J.~Berant, ``The web as a knowledge-base for answering complex
  questions,'' in \emph{{NAACL-HLT}}, 2018.

\bibitem{Han-EMNLP-2020}
J.~Han, B.~Cheng, and X.~Wang, ``Open domain question answering based on text
  enhanced knowledge graph with hyperedge infusion,'' in \emph{Findings of
  EMNLP}, 2020.

\bibitem{Apoorv-ACL-2020}
A.~Saxena, A.~Tripathi, and P.~P. Talukdar, ``Improving multi-hop question
  answering over knowledge graphs using knowledge base embeddings,'' in
  \emph{{ACL}}, 2020.

\bibitem{Zhang-arxiv-2022}
J.~Zhang, X.~Zhang, J.~Yu, J.~Tang, J.~Tang, C.~Li, and H.~Chen, ``Subgraph
  retrieval enhanced model for multi-hop knowledge base question answering,''
  \emph{CoRR}, vol. abs/2202.13296, 2022.

\bibitem{Han-IJCAI-2020}
J.~Han, B.~Cheng, and X.~Wang, ``Two-phase hypergraph based reasoning with
  dynamic relations for multi-hop kbqa,'' in \emph{{IJCAI}}, 2020.

\bibitem{Yasunaga-NAACL-2021}
M.~Yasunaga, H.~Ren, A.~Bosselut, P.~Liang, and J.~Leskovec, ``{QA-GNN:}
  reasoning with language models and knowledge graphs for question answering,''
  in \emph{{NAACL-HLT}}, 2021.

\bibitem{He-WSDM-2021}
G.~He, Y.~Lan, J.~Jiang, W.~X. Zhao, and J.~Wen, ``Improving multi-hop
  knowledge base question answering by learning intermediate supervision
  signals,'' in \emph{{WSDM}}, 2021.

\bibitem{Zhou-COLING-2018}
M.~Zhou, M.~Huang, and X.~Zhu, ``An interpretable reasoning network for
  multi-relation question answering,'' in \emph{{COLING}}, 2018.

\bibitem{Xu-NAACL-2019}
K.~Xu, Y.~Lai, Y.~Feng, and Z.~Wang, ``Enhancing key-value memory neural
  networks for knowledge based question answering,'' in \emph{{NAACL-HLT}},
  2019.

\bibitem{He-ACL-2017}
S.~He, C.~Liu, K.~Liu, and J.~Zhao, ``Generating natural answers by
  incorporating copying and retrieving mechanisms in sequence-to-sequence
  learning,'' in \emph{{ACL}}, 2017.

\bibitem{Yin-IJCAI-2016}
J.~Yin, X.~Jiang, Z.~Lu, L.~Shang, H.~Li, and X.~Li, ``Neural generative
  question answering,'' in \emph{{IJCAI}}, 2016.

\bibitem{Fu-NAACL-2018}
Y.~Fu and Y.~Feng, ``Natural answer generation with heterogeneous memory,'' in
  \emph{{NAACL-HLT}}, 2018.

\bibitem{Feng-EMNLP-2021}
Y.~Feng, J.~Zhang, G.~He, W.~X. Zhao, L.~Liu, Q.~Liu, C.~Li, and H.~Chen, ``A
  pretraining numerical reasoning model for ordinal constrained question
  answering on knowledge base,'' in \emph{Findings of {EMNLP}}, 2021.

\bibitem{Zhang-AAAI-2018}
Y.~Zhang, H.~Dai, Z.~Kozareva, A.~J. Smola, and L.~Song, ``Variational
  reasoning for question answering with knowledge graph,'' in \emph{{AAAI}},
  2018.

\bibitem{Min-NAACL-2013}
B.~Min, R.~Grishman, L.~Wan, C.~Wang, and D.~Gondek, ``Distant supervision for
  relation extraction with an incomplete knowledge base,'' in
  \emph{{NAACL-HLT}}, 2013.

\bibitem{Feng-AAAI-2019}
Y.~Feng, H.~You, Z.~Zhang, R.~Ji, and Y.~Gao, ``Hypergraph neural networks,''
  in \emph{{AAAI}}, 2019.

\bibitem{Trouillon-complexe-2016}
T.~Trouillon, J.~Welbl, S.~Riedel, {\'{E}}.~Gaussier, and G.~Bouchard,
  ``Complex embeddings for simple link prediction,'' in \emph{{ICML}}, 2016.

\bibitem{Liu-roberta-2019}
Y.~Liu, M.~Ott, N.~Goyal, J.~Du, M.~Joshi, D.~Chen, O.~Levy, M.~Lewis,
  L.~Zettlemoyer, and V.~Stoyanov, ``Roberta: {A} robustly optimized {BERT}
  pretraining approach,'' \emph{CoRR}, 2019.

\bibitem{ERNIE-ACL-2019}
Z.~Zhang, X.~Han, Z.~Liu, X.~Jiang, M.~Sun, and Q.~Liu, ``{ERNIE:} enhanced
  language representation with informative entities,'' in \emph{{ACL}}, 2019.

\bibitem{Peters-EMNLP-2019}
M.~E. Peters, M.~Neumann, R.~L.~L. IV, R.~Schwartz, V.~Joshi, S.~Singh, and
  N.~A. Smith, ``Knowledge enhanced contextual word representations,'' in
  \emph{{EMNLP-IJCNLP}}, 2019.

\bibitem{Foundation-arxiv-2021}
R.~Bommasani, D.~A. Hudson, E.~Adeli, R.~Altman, S.~Arora, S.~von Arx, M.~S.
  Bernstein, J.~Bohg, A.~Bosselut, E.~Brunskill, E.~Brynjolfsson, S.~Buch,
  D.~Card, R.~Castellon, N.~S. Chatterji, A.~S. Chen, K.~Creel, J.~Q. Davis,
  D.~Demszky, C.~Donahue, M.~Doumbouya, E.~Durmus, S.~Ermon, J.~Etchemendy,
  K.~Ethayarajh, L.~Fei{-}Fei, C.~Finn, T.~Gale, L.~Gillespie, K.~Goel, N.~D.
  Goodman, S.~Grossman, N.~Guha, T.~Hashimoto, P.~Henderson, J.~Hewitt, D.~E.
  Ho, J.~Hong, K.~Hsu, J.~Huang, T.~Icard, S.~Jain, D.~Jurafsky, P.~Kalluri,
  S.~Karamcheti, G.~Keeling, F.~Khani, O.~Khattab, P.~W. Koh, M.~S. Krass,
  R.~Krishna, R.~Kuditipudi, and et~al., ``On the opportunities and risks of
  foundation models,'' \emph{CoRR}, vol. abs/2108.07258, 2021.

\bibitem{Han-arxiv-2021}
X.~Han, Z.~Zhang, N.~Ding, Y.~Gu, X.~Liu, Y.~Huo, J.~Qiu, L.~Zhang, W.~Han,
  M.~Huang, Q.~Jin, Y.~Lan, Y.~Liu, Z.~Liu, Z.~Lu, X.~Qiu, R.~Song, J.~Tang,
  J.~Wen, J.~Yuan, W.~X. Zhao, and J.~Zhu, ``Pre-trained models: Past, present
  and future,'' \emph{CoRR}, vol. abs/2106.07139, 2021.

\bibitem{Yu-arxiv-2020}
W.~Yu, C.~Zhu, Z.~Li, Z.~Hu, Q.~Wang, H.~Ji, and M.~Jiang, ``A survey of
  knowledge-enhanced text generation,'' \emph{CoRR}, vol. abs/2010.04389, 2020.

\bibitem{Das-EMNLP-21}
R.~Das, M.~Zaheer, D.~Thai, A.~Godbole, E.~Perez, J.~Y. Lee, L.~Tan,
  L.~Polymenakos, and A.~McCallum, ``Case-based reasoning for natural language
  queries over knowledge bases,'' in \emph{{EMNLP}}, 2021.

\bibitem{Ye-ACL-22}
X.~Ye, S.~Yavuz, K.~Hashimoto, Y.~Zhou, and C.~Xiong, ``{RNG-KBQA:} generation
  augmented iterative ranking for knowledge base question answering,'' in
  \emph{{ACL}}, 2022.

\bibitem{KQA-Pro}
J.~Shi, S.~Cao, L.~Pan, Y.~Xiang, L.~Hou, J.~Li, H.~Zhang, and B.~He, ``Kqa
  pro: A large diagnostic dataset for complex question answering over knowledge
  base,'' \emph{CoRR}, 2020.

\bibitem{UnifiedSKG-arxiv-2022}
T.~Xie, C.~H. Wu, P.~Shi, R.~Zhong, T.~Scholak, M.~Yasunaga, C.~Wu, M.~Zhong,
  P.~Yin, S.~I. Wang, V.~Zhong, B.~Wang, C.~Li, C.~Boyle, A.~Ni, Z.~Yao, D.~R.
  Radev, C.~Xiong, L.~Kong, R.~Zhang, N.~A. Smith, L.~Zettlemoyer, and T.~Yu,
  ``Unifiedskg: Unifying and multi-tasking structured knowledge grounding with
  text-to-text language models,'' \emph{CoRR}, 2022.

\bibitem{Raffel-JMLR-2020}
C.~Raffel, N.~Shazeer, A.~Roberts, K.~Lee, S.~Narang, M.~Matena, Y.~Zhou,
  W.~Li, and P.~J. Liu, ``Exploring the limits of transfer learning with a
  unified text-to-text transformer,'' \emph{{JMLR}}, 2020.

\bibitem{PPR-Andersen-2006}
R.~Andersen, F.~R.~K. Chung, and K.~J. Lang, ``Local graph partitioning using
  pagerank vectors,'' in \emph{{FOCS}}, 2006.

\bibitem{Petroni-EMNLP-2019}
F.~Petroni, T.~Rockt{\"{a}}schel, S.~Riedel, P.~S.~H. Lewis, A.~Bakhtin, Y.~Wu,
  and A.~H. Miller, ``Language models as knowledge bases?'' in \emph{{EMNLP}},
  2019.

\bibitem{Bouraoui-AAAI-2020}
Z.~Bouraoui, J.~Camacho{-}Collados, and S.~Schockaert, ``Inducing relational
  knowledge from {BERT},'' in \emph{{AAAI}}, 2020.

\bibitem{Jiang-TACL-2020}
Z.~Jiang, F.~F. Xu, J.~Araki, and G.~Neubig, ``How can we know what language
  models know,'' \emph{{TACL}}, 2020.

\bibitem{Cao-ACL-2022}
S.~Cao, J.~Shi, Z.~Yao, L.~Hou, J.~Li, and J.~Xiao, ``Program transfer and
  ontology awareness for semantic parsing in {KBQA},'' \emph{CoRR}, 2021.

\bibitem{Zhang-KBS-2022}
J.~Zhang, L.~Zhang, B.~Hui, and L.~Tian, ``Improving complex knowledge base
  question answering via structural information learning,''
  \emph{Knowledge-Based Systems}, 2022.

\bibitem{Shi-EMNLP-2021}
J.~Shi, S.~Cao, L.~Hou, J.~Li, and H.~Zhang, ``Transfernet: An effective and
  transparent framework for multi-hop question answering over relation graph,''
  in \emph{{EMNLP}}, 2021.

\bibitem{Lan-ACL-2021}
Y.~Lan and J.~Jiang, ``Modeling transitions of focal entities for
  conversational knowledge base question answering,'' in \emph{{ACL}}, 2021.

\bibitem{Lan-IJCAI-2019}
Y.~Lan, S.~Wang, and J.~Jiang, ``Knowledge base question answering with topic
  units,'' in \emph{{IJCAI}}, 2019.

\bibitem{QALD}
Orange, ``Qald data challenge,'' \url{http://qald.aksw.org/}, 2011.

\bibitem{LC-QuAD-ISWC-2017}
P.~Trivedi, G.~Maheshwari, M.~Dubey, and J.~Lehmann, ``Lc-quad: {A} corpus for
  complex question answering over knowledge graphs,'' in \emph{{ISWC}}, 2017.

\bibitem{Zheng-CIKM-2020}
W.~Zheng and M.~Zhang, ``Automated query graph generation for querying
  knowledge graphs,'' in \emph{{CIKM}}, 2021.

\bibitem{Vakulenko-CIKM-2019}
S.~Vakulenko, J.~D.~F. Garcia, A.~Polleres, M.~de~Rijke, and M.~Cochez,
  ``Message passing for complex question answering over knowledge graphs,'' in
  \emph{{CIKM}}, 2019.

\bibitem{LC-QuAD-2.0-ISWC-2019}
M.~Dubey, D.~Banerjee, A.~Abdelkawi, and J.~Lehmann, ``Lc-quad 2.0: {A} large
  dataset for complex question answering over wikidata and dbpedia,'' in
  \emph{{ISWC}}, 2019.

\bibitem{Zou-arxiv-2021}
J.~Zou, M.~Yang, L.~Zhang, Y.~Xu, Q.~Pan, F.~Jiang, R.~Qin, S.~Wang, Y.~He,
  S.~Huang, and Z.~Zhao, ``A chinese multi-type complex questions answering
  dataset over wikidata,'' \emph{CoRR}, vol. abs/2111.06086, 2021.

\bibitem{Cai-ACL-2021}
J.~Cai, Z.~Zhang, F.~Wu, and J.~Wang, ``Deep cognitive reasoning network for
  multi-hop question answering over knowledge graphs,'' in \emph{{ACL/IJCNLP}},
  2021.

\bibitem{CFQ-ICLR-2020}
D.~Keysers, N.~Sch{\"{a}}rli, N.~Scales, H.~Buisman, D.~Furrer, S.~Kashubin,
  N.~Momchev, D.~Sinopalnikov, L.~Stafiniak, T.~Tihon, D.~Tsarkov, X.~Wang,
  M.~van Zee, and O.~Bousquet, ``Measuring compositional generalization: {A}
  comprehensive method on realistic data,'' in \emph{{ICLR}}, 2020.

\bibitem{Guo-NIPS-2020}
Y.~Guo, Z.~Lin, J.~Lou, and D.~Zhang, ``Hierarchical poset decoding for
  compositional generalization in language,'' in \emph{NeurIPS}, 2020.

\bibitem{Xiong-ACL-2022}
X.~Ye, S.~Yavuz, K.~Hashimoto, Y.~Zhou, and C.~Xiong, ``Rng-kbqa: Generation
  augmented iterative ranking for knowledge base question answering,''
  \emph{CoRR}, vol. abs/2109.08678, 2021.

\bibitem{Chen-ACL-2021}
S.~Chen, Q.~Liu, Z.~Yu, C.-Y. Lin, J.-G. Lou, and F.~Jiang, ``{R}e{T}ra{C}k: A
  flexible and efficient framework for knowledge base question answering,'' in
  \emph{ACL}, 2021.

\bibitem{Yao-EMNLP-2020}
Z.~Yao, Y.~Tang, W.~Yih, H.~Sun, and Y.~Su, ``An imitation game for learning
  semantic parsers from user interaction,'' in \emph{{EMNLP}}, 2020.

\bibitem{Zheng-IS-2019}
W.~Zheng, H.~Cheng, J.~X. Yu, L.~Zou, and K.~Zhao, ``Interactive natural
  language question answering over knowledge graphs,'' \emph{Inf. Sci.}, 2019.

\bibitem{Hua-EMNLP-2020}
Y.~Hua, Y.-F. Li, G.~Haffari, G.~Qi, and T.~Wu, ``Few-shot complex knowledge
  base question answering via meta reinforcement learning,'' in \emph{{EMNLP}},
  2020.

\bibitem{Zhou-NAACL-2021}
Y.~Zhou, X.~Geng, T.~Shen, W.~Zhang, and D.~Jiang, ``Improving zero-shot
  cross-lingual transfer for multilingual question answering over knowledge
  graph,'' in \emph{{NAACL-HLT}}, 2021.

\bibitem{Gardner-arXiv-2020}
M.~Gardner, Y.~Artzi, V.~Basmova, J.~Berant, B.~Bogin, S.~Chen, P.~Dasigi,
  D.~Dua, Y.~Elazar, A.~Gottumukkala, N.~Gupta, H.~Hajishirzi, G.~Ilharco,
  D.~Khashabi, K.~Lin, J.~Liu, N.~F. Liu, P.~Mulcaire, Q.~Ning, S.~Singh, N.~A.
  Smith, S.~Subramanian, R.~Tsarfaty, E.~Wallace, A.~Zhang, and B.~Zhou,
  ``Evaluating nlp models via contrast sets,'' \emph{ArXiv}, 2020.

\bibitem{Kaushik-ICLR-2020}
D.~Kaushik, E.~Hovy, and Z.~Lipton, ``Learning the difference that makes a
  difference with counterfactually-augmented data,'' in \emph{{ICLR}}, 2020.

\bibitem{Oren-EMNLP-2020}
I.~Oren, J.~Herzig, N.~Gupta, M.~Gardner, and J.~Berant, ``Improving
  compositional generalization in semantic parsing,'' in \emph{Findings of
  EMNLP}, 2020.

\bibitem{Guo-NIPS-2018}
D.~Guo, D.~Tang, N.~Duan, M.~Zhou, and J.~Yin, ``Dialog-to-action:
  Conversational question answering over a large-scale knowledge base,'' in
  \emph{{NeurIPS}}, 2018.

\bibitem{Christmann-CIKM-2019}
P.~Christmann, R.~S. Roy, A.~Abujabal, J.~Singh, and G.~Weikum, ``Look before
  you hop: Conversational question answering over knowledge graphs using
  judicious context expansion,'' in \emph{{CIKM}}, 2019.

\bibitem{Plepi-ESWC-2021}
J.~Plepi, E.~Kacupaj, K.~Singh, H.~Thakkar, and J.~Lehmann, ``Context
  transformer with stacked pointer networks for conversational question
  answering over knowledge graphs,'' in \emph{{ESWC}}, 2021.

\bibitem{kaiser:sigir2021}
M.~Kaiser, R.~S. Roy, and G.~Weikum, ``Reinforcement learning from
  reformulations in conversational question answering over knowledge graphs,''
  2021.

\bibitem{kacupaj:eacl2021}
E.~Kacupaj, J.~Plepi, K.~Singh, H.~Thakkar, J.~Lehmann, and M.~Maleshkova,
  ``Conversational question answering over knowledge graphs with transformer
  and graph attention networks,'' in \emph{{EACL}}, 2021.

\bibitem{Vaswani-NIPS-2017}
A.~Vaswani, N.~Shazeer, N.~Parmar, J.~Uszkoreit, L.~Jones, A.~N. Gomez,
  L.~Kaiser, and I.~Polosukhin, ``Attention is all you need,'' in
  \emph{{NuerIPS}}, 2017.

\bibitem{Lamb-IJCAI-2020}
L.~C. Lamb, A.~S. d'Avila Garcez, M.~Gori, M.~O.~R. Prates, P.~H.~C. Avelar,
  and M.~Y. Vardi, ``Graph neural networks meet neural-symbolic computing: {A}
  survey and perspective,'' in \emph{{IJCAI}}, 2020.

\bibitem{Arabshahi-ICLR-2018}
F.~Arabshahi, S.~Singh, and A.~Anandkumar, ``Combining symbolic expressions and
  black-box function evaluations in neural programs,'' in \emph{{ICLR}}.\hskip
  1em plus 0.5em minus 0.4em\relax OpenReview.net, 2018.

\bibitem{Lample-ICLR-2020}
G.~Lample and F.~Charton, ``Deep learning for symbolic mathematics,'' in
  \emph{{ICLR}}.\hskip 1em plus 0.5em minus 0.4em\relax OpenReview.net, 2020.

\bibitem{Xie-AAAI-2016}
R.~Xie, Z.~Liu, J.~Jia, H.~Luan, and M.~Sun, ``Representation learning of
  knowledge graphs with entity descriptions,'' in \emph{{AAAI}}, 2016.

\bibitem{Xie-IJCAI-2017}
R.~Xie, Z.~Liu, H.~Luan, and M.~Sun, ``Image-embodied knowledge representation
  learning,'' in \emph{{IJCAI}}, 2017.

\bibitem{He-WWW-2020}
G.~He, J.~Li, W.~X. Zhao, P.~Liu, and J.~Wen, ``Mining implicit entity
  preference from user-item interaction data for knowledge graph completion via
  adversarial learning,'' in \emph{{WWW}}, 2020.

\bibitem{Lu-SIGIR-2019}
X.~Lu, S.~Pramanik, R.~S. Roy, A.~Abujabal, Y.~Wang, and G.~Weikum, ``Answering
  complex questions by joining multi-document evidence with quasi knowledge
  graphs,'' in \emph{{SIGIR}}, 2019.

\bibitem{Dhingra-ICLR-2020}
B.~Dhingra, M.~Zaheer, V.~Balachandran, G.~Neubig, R.~Salakhutdinov, and W.~W.
  Cohen, ``Differentiable reasoning over a virtual knowledge base,'' in
  \emph{{ICLR}}, 2020.

\end{thebibliography}


\begin{thebibliography}{1}

\bibitem{IEEEhowto:kopka}
H.~Kopka and P.~W. Daly, \emph{A Guide to \LaTeX}, 3rd~ed.\hskip 1em plus
  0.5em minus 0.4em\relax Harlow, England: Addison-Wesley, 1999.

\end{thebibliography}
